\documentclass[10pt,reqno]{amsart}
\usepackage{hyperref}
\usepackage{latexsym,amsmath}
\usepackage{enumerate}
\usepackage{amsfonts}
\usepackage{amssymb}
\usepackage{latexsym}
\usepackage{fixmath}
\usepackage[usenames,dvipsnames]{color}

\usepackage{graphicx}
\usepackage{multirow}

\usepackage[T1]{fontenc}
\usepackage{fourier}
\usepackage{bbm}
\usepackage{color}
\usepackage{fullpage}

\numberwithin{equation}{section}

\newcommand{\DD}{\mathtt{DD}}

\newcommand{\vsigma}{\SIGMA}
\newcommand{\hvsigma}{\hat\SIGMA}

\renewcommand{\vec}[1]{\boldsymbol{#1}}

\newcommand\vk{\vec k}

\newcommand\nix{\,\cdot\,}

\newcommand\vA{\vec A}

\newcommand\G{\vec G}

\newcommand\SIGMA{\vec\sigma}

\newcommand\cA{\mathcal{A}}
\newcommand\cB{\mathcal{B}}

\newcommand\cD{\mathcal{D}}

\newcommand\cH{\mathcal{H}}

\newcommand\cM{\mathcal{M}}

\newcommand\cX{\mathcal{X}}

\newcommand{\vecone}{\vec{1}}

\newcommand{\Bin}{{\rm Bin}}

\newcommand{\Be}{{\rm Be}}

\newcommand\bc[1]{\left({#1}\right)}
\newcommand\cbc[1]{\left\{{#1}\right\}}

\newcommand{\bck}[1]{\left\langle{#1}\right\rangle}
\newcommand\brk[1]{\left\lbrack{#1}\right\rbrack}

\newcommand\RR{\mathbb{R}}

\newcommand\pr{\mathbb{P}}

\newcommand\Sec{Section}

\DeclareMathOperator{\rank}{rk}

\def\pr{{\mathbb P}}

\def\cH{{\mathcal H}}

\newcommand{\abp}{adaptive Belief Propagation}

\begin{document}
	
\title{Efficient and accurate group testing via Belief Propagation: an empirical study}

\author{Amin~Coja-Oghlan, Max Hahn-Klimroth, Philipp Loick, Manuel Penschuck}

\thanks{Amin Coja-Oghlan, Max Hahn-Klimroth and Philipp Loick are supported by DFG CO 646/3 and DFG CO 646/5. Manuel Penschuck is supported by ME 2088/5-1.}

\address{Amin Coja-Oghlan, {\tt acoghlan@math.uni-frankfurt.de}, Goethe University, Mathematics Institute, 10 Robert Mayer St, Frankfurt 60325, Germany.}
\address{Max Hahn-Klimroth, {\tt hahnklim@math.uni-frankfurt.de}, Goethe University, Mathematics Institute, 10 Robert Mayer St, Frankfurt 60325, Germany.}
\address{Philipp Loick, {\tt loick@math.uni-frankfurt.de}, Goethe University, Mathematics Institute, 10 Robert Mayer St, Frankfurt 60325, Germany.}
\address{Manuel Penschuck, {\tt mpenschuck@ae.cs.uni-frankfurt.de}, Goethe University, Institute for Computer Science, 11--15 Robert Mayer St, Frankfurt 60325, Germany.}

\begin{abstract}
	The group testing problem asks for efficient pooling schemes and algorithms that allow to screen moderately large numbers of samples for rare infections.
	The goal is to accurately identify the infected samples while conducting the least possible number of tests.
	Exploring the use of techniques centred around the Belief Propagation message passing algorithm, we suggest a new test design that significantly increases the accuracy of the results.
	The new design comes with Belief Propagation as an efficient inference algorithm.
	Aiming for results on practical rather than asymptotic problem sizes, we conduct an experimental study. \textit{MSc: 05C80, 60B20, 68P30}
\end{abstract}

\maketitle

\section{Introduction}\label{Sec_intro}

\subsection{The group testing problem}\label{Sec_problem}
In science generally and in applied science particularly there is much to be said for simplicity.
But occasionally a modest degree of sophistication carries extraordinary rewards.
Group testing is a case in point.
Every single day medical labs around the globe screen moderately large numbers of samples for rare pathogens.
The vast majority of samples, anywhere between 90\%\ and 99.9\%, are actually uninfected \cite{brault_2021, kleinman_2005, mallapaty_2020, ohhashi_2010, shental_2020, sherlock_2007, tebbs_2013, theagarajan_2020, vanZyl_2011}.
Labs therefore test pools of samples rather than individual samples.
The {\em group testing problem} asks for pooling strategies that minimise the total number of tests required while maximising the accuracy of the results.
The latter is crucial because test results are generally not perfectly accurate.

Coming up with practical solutions to this problem turns out to be challenging precisely because the total number of samples in a real-world scenario is {moderate}--say, in the hundreds or thousands.
To elaborate, on the one hand the group testing problem has inspired a body of beautiful mathematical work that deals with the asymptotical scenario where the number of samples grows to infinity \cite{Aldridge_2019, Coja_2019, COLT}.
However, such asymptotical results do not directly bear on practical problem sizes.
Besides, the theoretical test designs tend to suffer other drawbacks such as asking for excessively large test pools or subdivisions of individual samples into very many sub-samples \cite{Aldridge_2019, Coja_2019, COLT}.
On the other hand, practical problem sizes far exceed the limits up to which an exhaustive search for an optimal test design seems remotely feasible.
As a consequence, the pooling schemes in practical use remain the self-same extremely simple ones suggested in the 1940s \cite{brault_2021, kleinman_2005, mallapaty_2020, ohhashi_2010, shental_2020, sherlock_2007, tebbs_2013, theagarajan_2020, vanZyl_2011}. 

The aim of this paper is to investigate better test designs for practical problem sizes.
The focus is on improving the {\em accuracy} of the results, i.e., avoiding false positive and/or negative diagnoses while keeping the number of tests as small as possible.
Indeed, the thrust of this paper is that the idea of group testing, originally invented to reduce the number of tests, actually excels at improving the accuracy of the results.
This may seem surprising at first glance because one might deem individual testing optimal in terms of accuracy.
It is not.
Group testing does better in much the same way as error-correcting codes gain power from encoding entire blocks of data simultaneously.

Given the moderate number of samples in real-world scenarios, an empirical approach is the only feasible way to obtain practically meaningful results.
Thus, taking on board the intuition from theoretical work on group testing as well as recent ideas from information theory and statistical physics, we conduct an extensive experimental study.
The main finding is that a novel test scheme called {\em \abp} greatly improves the accuracy of the overall results while keeping the number of tests conducted low.
Furthermore, the new test design requires only relatively small test pools and only assigns each sample to a small number of tests.
Finally, the design comes with an efficient, easy-to-implement algorithm to infer the status of the individual samples from the test results, namely the Belief Propagation message passing algorithm.

We proceed to discuss the mathematical model of tests and samples that we work with.
Subsequently, we present the results of \abp\ by comparison to other test schemes.
These test schemes, which are partly incorporated into \abp, are discussed in detail in \Sec~\ref{sec_known}.
In \Sec~\ref{sec_adabp} we then present the new test design and the corresponding inference algorithm.
\Sec~\ref{sec_asymptotic} details the theoretical and heuristical considerations that underpin \abp.
Finally, in \Sec~\ref{sec_discussion} we discuss the potential impact of the new results and future directions for both empirical and theoretical work.

\subsection{The model}\label{Sec_model}
We work with a simple but standard model of group testing that allows for test results to not be entirely accurate \cite{Aldridge_2019}.
Let $x_1,\ldots,x_n$ represent the samples submitted for testing.
We assume that with a prior probability of $\lambda\in[0,1]$ any one sample is infected is known.
The true infection status of each sample is indicated by $\vsigma(x_j)\in\{0,1\}$, with $1$ representing `infected'. 
The $\vsigma(x_j)$ are assumed to be independent Bernoulli variables with mean $\lambda$. 
We refer to the vector $\vsigma=(\vsigma(x_j))_{j=1,\ldots,n}$ as the {\em ground truth}.
Let $\vk=\sum_{j=1}^n\vecone\{\vsigma(x_j)=1\}$ signify the actual number of infected samples.

The way how test pools are formed is represented by a bipartite graph.
To be precise, a \emph{test design} is a bipartite graph $G$ with one class $\cX=\{x_1,\ldots,x_n\}$ of vertices representing the $n$ samples and the other class $\cA=\{a_1,\ldots,a_m\}$ representing the test pools.
An edge between $x_j$ and $a_i$ indicates that $x_j$ is included in test pool $a_i$.
For each $x_j$ we let $\partial x_j=\partial_G x_j$ be the set of test pools that include $x_j$.
Similarly, for each test pool $a_i$ we write $\partial a_i$ for the set of individual samples $x_j$ included in that pool.

Each test $a_i$ reports a positive or negative result $\hvsigma(a_i)\in\{0,1\}$.
Ideally a test $a_i$ should come back positive iff at least one sample $x_j\in\partial a_i$ is actually infected.
But the test results need not be completely accurate.
We therefore posit two parameters $p$, called the {\em specificity},  
and $q$, the {\em sensitivity}, both between $0$ and $1$, 
such that the tests return results
\begin{align}\label{eqActuallyNeg}
	\hvsigma(a_i)&=\begin{cases}0&\mbox{ with probability }p\\ 1&\mbox{ with probability }1-p\end{cases}&&\mbox{if }\vsigma(x_j)=0\mbox{ for all }x_j\in\partial a_i\\
	\hvsigma(a_i)&=\begin{cases}0&\mbox{ with probability }1-q\\1&\mbox{ with probability }q\end{cases}&&\mbox{if }\vsigma(x_j)=1\mbox{ for some }x_j\in\partial a_i.\label{eqActuallyPos}
\end{align}
The random outcomes in~\eqref{eqActuallyNeg}--\eqref{eqActuallyPos} are mutually independent given $\vsigma$.
Let $\hvsigma=(\hvsigma(a_i))_{i=1,\ldots,m}$ encompass the test results.

Generally the ground truth $\vsigma$ cannot be inferred with perfect accuracy form the test results $\hvsigma$ of a single ``one-shot'' test design (unless $p=q=1$ and we test every $x_j$ separately) \cite{aldridge_2018}.
Indeed, under the noise model \eqref{eqActuallyNeg}--\eqref{eqActuallyPos} the posterior of the ground truth given the test results reads%
	\footnote{In \eqref{eqNishi} the $\propto$-symbol hides the normalisation required to turn $\mu_G$ into a probability distribution.}
\begin{align}\label{eqNishi}
	\mu_G(\sigma)=\pr\brk{\vsigma=\sigma\mid\hvsigma}&\propto \prod_{i=1}^n\lambda^{\sigma(x_i)}(1-\lambda)^{1-\sigma(x_i)}\prod_{i=1}^m \psi_{\hvsigma(a_i)}\bc{(\sigma(y))_{y\in \partial a_i}}\qquad(\sigma=(\sigma(x_i))_{i=1,\ldots,n}\in\{0,1\}^n) \\
	\mbox{where}\quad&\psi_0(\sigma_1,\ldots,\sigma_\ell)=
	p^{1-\bigvee_{i=1}^\ell\sigma_i}(1-q)^{\bigvee_{i=1}^\ell\sigma_i},\qquad
	\psi_1(\sigma_1,\ldots,\sigma_\ell)=
	(1-p)^{1-\bigvee_{i=1}^\ell\sigma_i}q^{\bigvee_{i=1}^\ell\sigma_i}.
\end{align}
Hence, the information-theoretically optimal inference algorithm just draws a random sample from the distribution $\mu_G$.
In effect, the accuracy with which the ground truth can potentially be recovered is governed by the entropy of the posterior $\mu_G$: the smaller the entropy the better the results.
Furthermore, depending on the specific design $G$ there may or may not exists an {\em efficient} algorithm for sampling from $\mu_G$.

To deal with these challenges, in {\em adaptive group testing} tests are not deployed in a single stage like in \eqref{eqActuallyNeg}--\eqref{eqActuallyPos} but in several stages.
To be precise, an {\em $\ell$-stage test design} is an increasing sequence $G^{(0)},G^{(1)},\ldots,G^{(\ell)}$ of test designs such that $G^{(i+1)}$ is obtained from $G^{(i)}$ by adding further tests.
How many tests are added and which samples they contain depends on results from the previous stages.
The results of the new tests are assumed to be distributed independently according to \eqref{eqActuallyNeg}--\eqref{eqActuallyPos}.
The aim, of course, is to diligently add tests so as to maximally reduce the entropy of the posterior.

In summary, the group testing problem poses the following challenges.
\begin{enumerate}[(i)]
	\item To come up with an adaptive test design that allows to infer the true infection status $\vsigma(x_j)$ of as many $x_j$ as possible while conducting as small a number of tests as possible. 
	\item To devise an {\em efficient} algorithm that actually infers the $\vsigma(x_j)$ from the observed $\hvsigma(a_i)$ with reasonable computational effort.
	\item To facilitate practical adoption we need to avoid high degrees because very large test pools may be infeasible, as may be dividing an individual sample into very many pools. 
	\item To ensure a timely reporting of test outcomes we should aim for a small number of test stages, or at least ensure that most samples can be diagnosed after the first or second stage.
\end{enumerate}

\subsection{Results}\label{Sec_results}
To meet these objectives we devise a new test scheme called \abp.
We investigate its performance empirically for the following parameter choices.
\begin{itemize}
	\item The results in this section refer to $n=1000$ samples.
		In \Sec~\ref{sec_scale} we discuss that the performance is similar on instances with $n=100$ and slightly better with $n=10000$.
	\item We study prior infection probabilities $ \lambda=0.5\%,\ 1\%,\ 5\%,\ 10\%$.
	\item Three different specificity/sensitivity scenarios are investigated:
		\begin{enumerate}[(a)]
\item perfectly reliable tests, i.e. $p=q=1$,
	\item moderately high values $p=0.99$ and $q=0.98$ which reflects, among others, the reliability of certain Covid-19 tests \cite{brault_2021, cohen_2020, mueller_2020, theagarajan_2020,whatson_2020} and 
	\item a noisy scenario with $p=q=0.95$.
\end{enumerate}
	\item Each experiment is run $100$ times independently for each parameter combination.
\end{itemize}

\begin{figure}[h]
	\centering
	\includegraphics[scale=0.32]{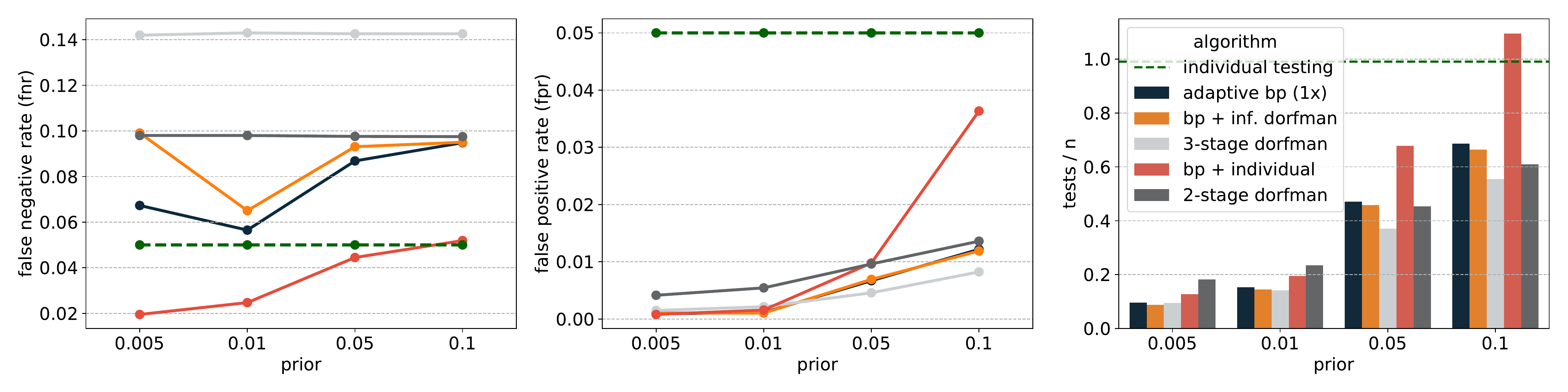}
	\caption{Simulation results for high noise scenario (sensitivity and specificity of $95\%$)}
	\label{fig_simulation_noise_high}
\end{figure}

The experiments show that \abp\ improves the accuracy of the results by an order of magnitude by comparison to known test designs while keeping the number of tests at a reasonable level.
Let us begin with the high-noise scenario $p=q=0.95$, where we reap the greatest gains.
We propose three different test designs \abp~1, \abp~2 and \abp~3.
The first strikes a balance between accuracy of results and the number of tests, while the latter emphasises accuracy.
In the following, let the \textit{false positive rate (fpr)} be the number of healthy samples falsely classified as infected over all healthy samples. Correspondingly, let the \textit{false negative rate (fnr)} be the number of infected samples falsely classified as healthy over all infected samples.
Figure~\ref{fig_simulation_noise_high} displays the results of \abp~1 in comparison to several previously known approaches.
These include the {\em 2-stage Dorfman} and the {3-stage Dorfman} designs, which are widely used in practice, as well as {\em Belief Propagation} followed by individual testing advocated in the theoretical literature \footnote{Note that with perfectly reliable tests, this approach is equivalent to the so-called {\em definite defectives (DD)} algorithm followed by individual testing.}.
A further scheme that we included is {\em informative Dorfman}, a 2-stage design proposed in~\cite{mcmahan_2012}.
We will discuss these approaches in some detail in \Sec~\ref{sec_known}.
Figure~\ref{fig_simulation_noise_high} shows that with about the same number of tests as 2-stage Dorfman, \abp\ achieves up to $78\%$ reduction in the number of false positives and an up to $42\%$ reduction in the number of false negatives. 
The gains are particularly high for small priors.

\begin{figure}[h]
	\centering
	\includegraphics[scale=0.32]{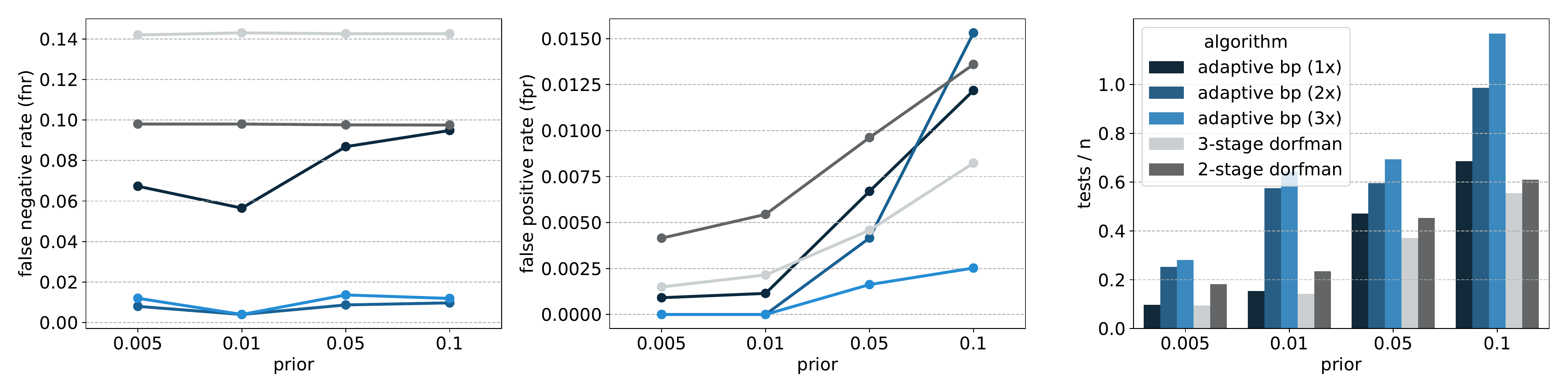}
	\caption{Simulation results of reliability-enhanced \abp\ for high noise scenario (sensitivity and specificity of $95\%$)}
	\label{fig_simulation_noise_high_triple}
\end{figure}

Nevertheless, the absolute value of the false positive and false negative rate of all test designs in Figure~\ref{fig_simulation_noise_high}, particularly for large priors, may still be unacceptably high for many real-world applications.
Here our other two designs \abp~2 and \abp~3 come to the rescue.
As Figure~\ref{fig_simulation_noise_high_triple} shows, these designs, particularly \abp~3, dramatically reduce the number of false positives and negatives.
Of course, these improvements come at the expense of a larger number of tests.
But for priors $\lambda\leq0.05$ the number of extra tests is moderate, and for the largest prior $\lambda=0.1$ \abp~2 and \abp~3 require not many more tests than individual testing while being the only designs that deliver decent accuracy.

\begin{figure}[h]
    \centering
    \includegraphics[scale=0.32]{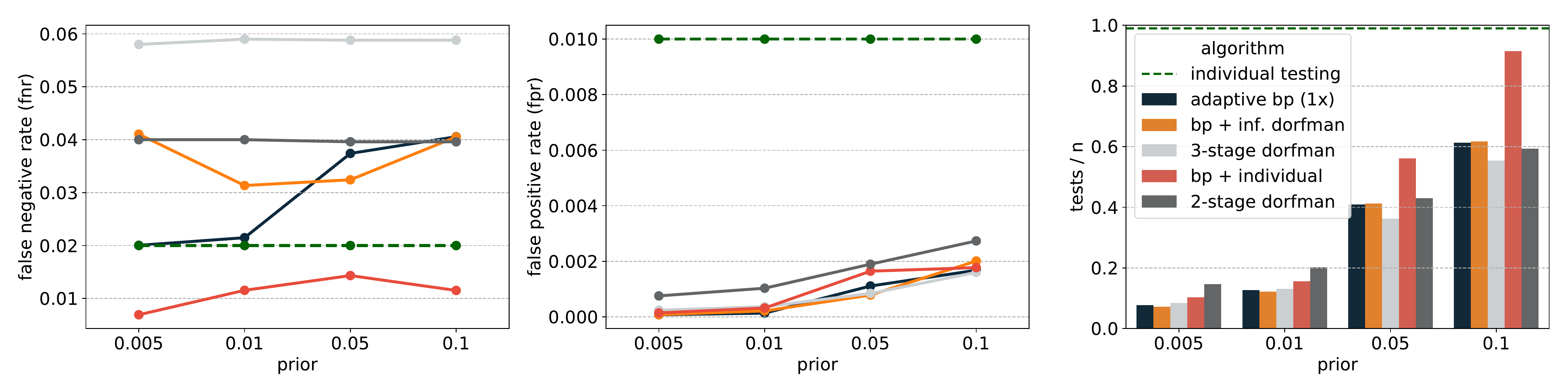}
    \caption{Simulation results for sensitivity for moderate noise scenario (sensitivity of $99\%$, specificity of $98\%$)}
    \label{fig_simulation_noise_low}
\end{figure}

\begin{figure}[h]
	\centering
	\includegraphics[scale=0.32]{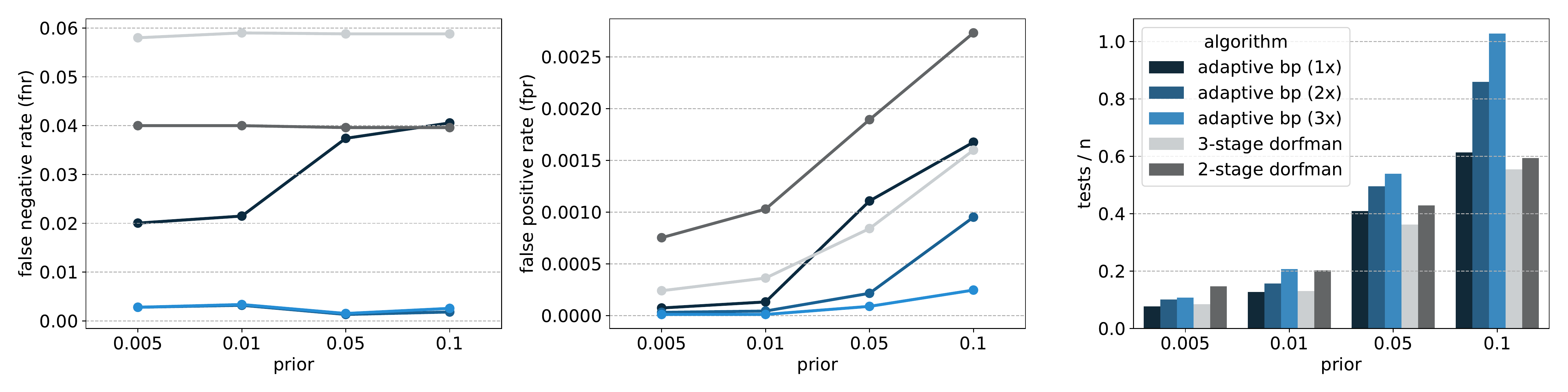}
	\caption{Simulation results of reliability-enhanced \abp\ for moderate noise scenario (sensitivity of $99\%$, specificity of $98\%$)}
	\label{fig_simulation_noise_low_triple}
\end{figure}

Matters turns out similar in the case of moderately high sensitivity and specificity $p=0.99$, $q=0.98$.
Figure~\ref{fig_simulation_noise_low} displays the results.
In comparison to the classical two- and three-stage Dorfman scheme, \abp\ requires at most $11\%$ more tests for high priors of $\lambda=0.1$ - for small priors even fewer tests. 
The benefit is that \abp\ boosts accuracy compared to all the previously known designs, particularly so for low priors. We point out that the gains vis-a-vis {\em informative Dorfman} for moderately high priors are modest. The key benefit in \abp\, however, lies in its versatility to meaningfully enhance the accuracy at the expense of somewhat more tests as shown in Figure~\ref{fig_simulation_noise_low_triple}. A similar extension of {\em informative Dorfman} would yield a similar accuracy but require significantly more tests than \abp.

Even with perfectly reliable tests, the conventional {\em definite defectives (DD)} approach in the literature can be improved upon by \abp\ or the {\em informative Dorfman} approach. Both schemes are able to reduce the number of tests compared to the former by up to $18\%$ and comes within $19\%$ to $32\%$ of the information-theoretic lower bound. The gains vis-a-vis two-stage Dorfman with up to $57\%$ and individual testing with up to $94\%$ are even more pronounced. We do not need to consider the accuracy in the noiseless case since all test designs recover the entire ground truth by construction.

\begin{figure}[h]
	\centering
	\includegraphics[scale=0.4]{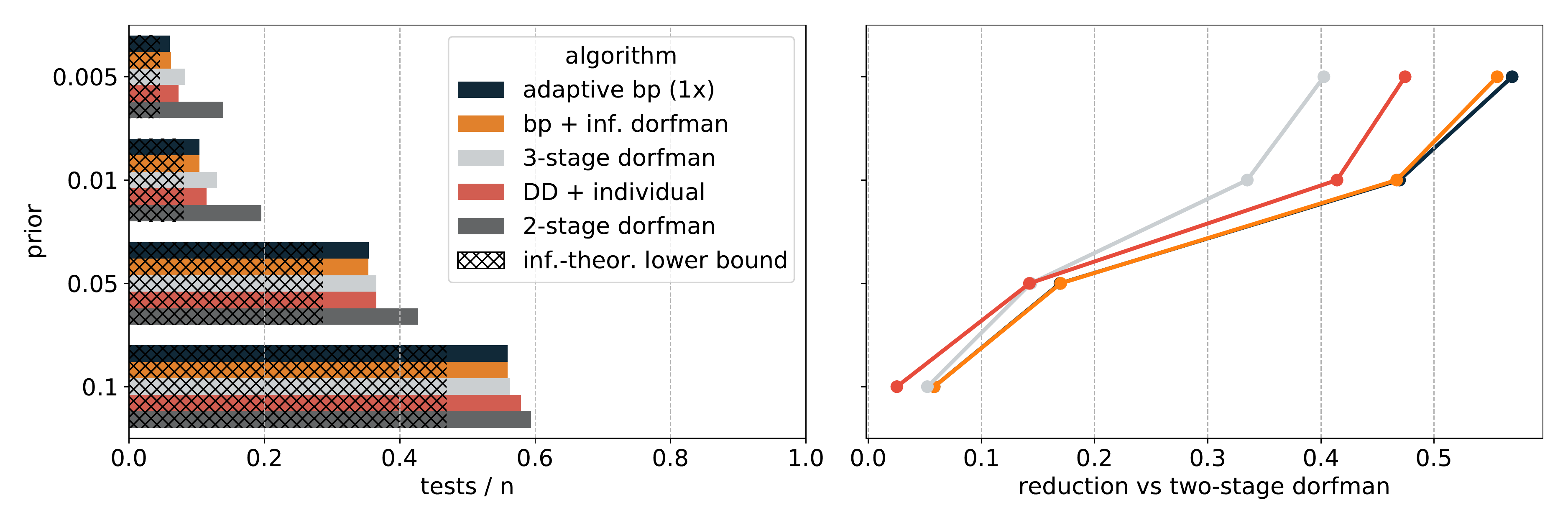}
	\caption{Simulation results for the noiseless setting. The left plot displays the numbers of tests required by the different designs; the black hatched area represents a plausible information-theoretic lower bound for the number of tests. The right plot shows the reduction achieved by comparison to the 2-stage Dorfman procedure, a classical and widely used test design.}
	\label{fig_simulation_noise_no}
\end{figure}

All examined algorithms require reasonable pool sizes and splits of the individual sample that are in line with common pooling procedures \cite{garrison_2015, joly_2020, mcmahan_2012}. 
The maximum pool size is between $8$ and $170$ depending on noise level and prior, while the splits of the individual sample range between $3$ and $19$. 
It should be noted that the proposed algorithms and test designs can readily be adjusted to accommodate smaller pool sizes or individual sample splits --- at the expense of somewhat more tests.

\subsection*{Organisation}

In \Sec~\ref{sec_known}, we will discuss designs and algorithms that are in practical use or have been studied in the mathematical literature on group testing. 
Subsequently, we present the details behind our novel test design named \abp\ in \Sec~\ref{sec_adabp}. 
In \Sec~\ref{sec_asymptotic} we relate \abp\ to the theoretical work on group testing and asymptotic considerations.

\section{Designs and algorithms}\label{sec_known}

\noindent
We discuss the various previously studied test designs and inference algorithms.
In \Sec~\ref{sec_adabp} we will then see how we extend and modify these known constructions in order to obtain the new \abp\ design.

\subsection{Individual testing}\label{Sec_ind}
The most straightforward test strategy, of course, is to conduct $m=n$ individual tests for each of the $n$ samples.
At first glance, individual testing may appear to be the gold standard in terms of accuracy.
Naturally, in the case $p=q=1$, individual testing will register the status of each sample correctly.
However, realistic values for $p$ and $q$ range between $0.95$ and $0.99$ \cite{brault_2021, cohen_2020, mueller_2020, theagarajan_2020,whatson_2020}.
If $p,q$ are less than one, then individual testing will produce numbers of false positives/negatives distributed as $\Bin(n-\vk,1-p)$ and $\Bin(\vk,1-q)$, respectively.

The accuracy of the results could obviously be boosted by conducting two or three individual tests per sample.
Indeed, if we test each $x_j$ twice and report $x_j$ as infected only if both tests come back positive, then we could reduce the expected number of false positives to $(n-\vk)(1-p)^2$.
But we would now expect a slightly larger number of $2\vk(1-q)$ false negatives.
To reduce the number of false positives and negatives simultaneously we could test each $x_j$ thrice and report the majority of the three test results.

\subsection{Dorfman and grid designs}
\begin{figure}[h]
	\centering
	\begin{minipage}[c]{0.6\textwidth}
		\centering
		\includegraphics[scale=0.9]{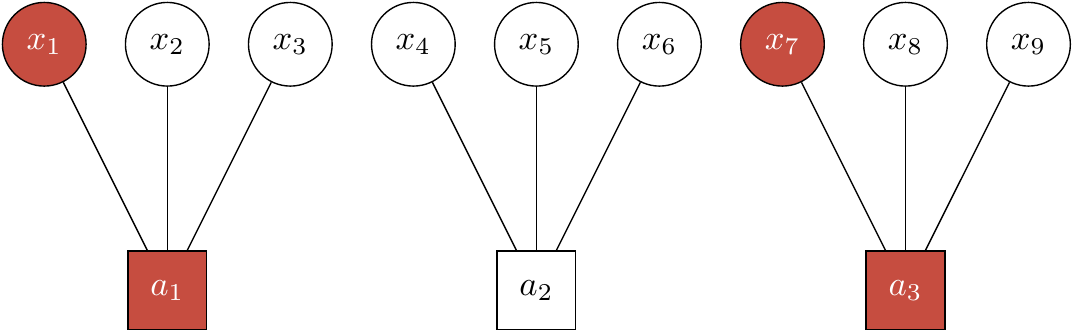}
	\end{minipage}
	\begin{minipage}[c]{0.39\textwidth}
		\centering
		\includegraphics[scale=0.9]{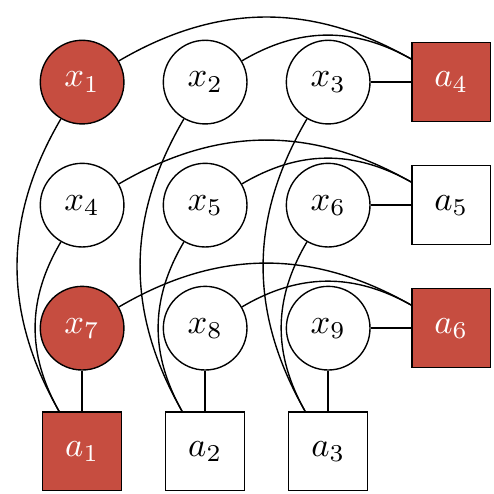}
	\end{minipage}
	\caption{Illustration of first stage of the first stage of the Dorfman scheme (left) and grid test designs (right)}
	\label{fig_dorf_grid_design}
\end{figure}

The test designs that currently appear to be most widely used in practice date back to the 1940s.
Indeed, the idea of group testing was first brought up by Dorfman in 1943 \cite{Dorfman_1943}.
He suggested a two-stage test procedure.
In the first stage, every sample gets placed in precisely one pool.
All pools are the same size, which depends on the prior $\lambda$ only.
Pools with a positive test result get tested separately in the second stage. 
An illustration is provided in Figure~\ref{fig_dorf_grid_design}.
Depending on the prior, this scheme can significantly reduce the number of tests required.
For example, with $\lambda=0.05$ this scheme uses pools of size five and the expected overall number of tests conducted in both stages comes to about $0.426n$.
At the same time, Dorfman's two-stage procedure reduces the number of false positives because a sample is ultimately reported as positive only if both the tests are positive.
But for the same reason, the expected number of false negatives increases.
For instance, with $n=10^4$ and $k=\lambda n=500$ as above, we expect $18.2$ false positives and $9.95$ false negatives.

A natural extension of the Dorfman procedure employs three stages.
In the first stage, relatively large pools are formed.
The second stage then splits the positive ones into smaller sub-pools and the third stage resorts to individual testing.
In effect, as with the two-stage procedure, the expected number of false positives drops while the expected number of false negatives increases.
For $n=10^4$ and $\lambda=0.05$ as above the expected numbers of false positives/negatives work out to be $11.76$ and $14.8$, respectively.

Grid designs are a variation on the Dorfman scheme.
They partition the set of all individual samples into equal-sized subsets.
For instance, if $\lambda=0.05$ the size would be 16.
Each of these subsets is mapped onto a $4\times 4$ grid.
Its rows and columns constitute the pools for the first stage.
Thus, each sample lands in two first-stage pools.
Depending on the results, further tests are conducted in a second stage; see Figure~\ref{fig_dorf_grid_design} for an illustration.
Grid designs significantly reduce the number of false negatives by comparison to individual testing while increasing the number of false positives.
However, the number of tests required exceeds that of the two-stage Dorfman procedure.

\subsection{Probabilistic constructions}\label{Sec_bireg}

\begin{figure}[ht]
	\centering
	\includegraphics[scale=0.9]{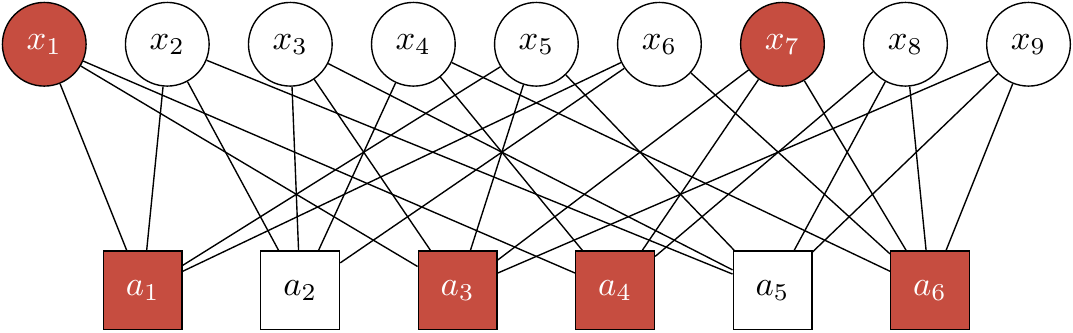}
	\caption{Illustration of a random biregular test design with $\Delta=3$ and $\Gamma=4$}
	\label{fig_regular_design}
\end{figure}

More sophisticated test designs have been proposed in the mathematical theory of group testing.
The best current, and in certain asymptotic settings provably optimal, test designs harness randomisation \cite{Aldridge_2019, COLT}.
For instance, in the {\em random biregular test design} every test pool has an equal size $\Gamma$ and every individual sample joins an equal number $\Delta$ of pools; see Figure~\ref{fig_regular_design} for an illustration.
In other words, the test design $G=G_{n,m}(\Gamma,\Delta)$ is chosen uniformly at random from the set of all $(\Delta,\Gamma)$-regular bipartite graphs \cite{Frieze2015}.%
\footnote{To be precise, $G$ is typically drawn from the pairing model of graphs with given degrees. In this model, it is rare but possible that the same individual joins a test pool twice. In practice, such double occurrence could, of course, be reduced to single occurrences.}
In order to extract the maximum amount of information about the ground truth, the parameters $\Gamma,\Delta$ should be chosen so as to maximise the conditional entropy of the vector $\hvsigma$ of test results.
Hence, $\Gamma,\Delta$ should be chosen so that about half the tests will be positive:%
\footnote{Of course, due to rounding issues we cannot ensure that the expected number of positive tests is precisely equal to $m/2$.}
\begin{align}\label{eqGammaDelta}
	\Delta&=\frac{m \log(2)}{n \lambda}&\Gamma&=\frac{\log(2)}{\lambda}.
\end{align}

Why does such a randomised construction seem promising?
Intuitively the randomness of the test design reduces dependencies between the different test results $\hvsigma(a_i)$ to a minimum.
Thus, with $\Delta,\Gamma$ chosen as above and for a number $m$ of tests up to a certain threshold, we can hope to squeeze as much as one bit's worth of information from each test.
Similar randomised constructions have proved powerful in coding theory and compressed sensing as well~\cite{Donoho_2013, Donoho_2006, Krzakala_2012,RichardsonUrbanke}.

While in the designs that we discussed previously obvious inference algorithms suggested themselves, in the case of the random biregular design matters are not quite so straightforward.
In the case $p=q=1$ maximum a posteriori inference boils down to a minimum hypergraph vertex cover problem~\cite{Coja_2019}.
However, this problem is NP-hard and even on random instances no efficient algorithm is known.

A blunt but efficient algorithm that has been analysed in the case $p=q=1$ goes by the name of {\em definite defectives} (`$\DD$')~\cite{aldridge_baldassini_2014}.
The algorithm classifies as infected every sample that is included in positive test pools only and that appears in at least one positive test pool where all other samples appear in a negative test.
All other samples are classified as uninfected.
In symbols,
\begin{align*}
	\vsigma_{\DD}(x_j)&=\bigwedge_{a\in\partial x_j}\hvsigma(a)\wedge\bigvee_{a\in\partial x_j}\bigwedge_{\substack{y\in\partial a\\y\neq x_j}}\bigvee_{b\in\partial y}(1-\hvsigma(b)).
\end{align*}
For $p=q=1$ this algorithm will never produce false positives but may render false negatives.
Several similarly-flavoured algorithms have been analysed mathematically.
Aldridge analysed an adaptive test design whose different stages employ random biregular test designs with suitably chosen degrees \cite{aldridge_2020}. 
This adaptive test design carried out over an unbounded number of stages achieves rates in excess of $0.95$ bits per tests. 
However, the large number of stages might render the scheme impractical.

\subsection{Glauber dynamics}\label{Sec_Glauber}

\noindent
The $\DD$ algorithm merely extracts binary information about each sample.
For a more fine-grained picture we would need to get a better handle on the posterior distribution \eqref{eqNishi} of the random test design.
An immediate idea is to use a Markov Chain Monte Carlo algorithm to approximate the marginals of the posterior.
Specifically, the Glauber dynamics starts at a random initial configuration $\sigma^{(0)}=(\sigma^{(0)}(x_i))_{i=1,\ldots,n}$ drawn from the prior.
Thus, the individual $\sigma^{(0)}(x_i)$ are independent $\Be(\lambda)$ variables.
Glauber then proceeds to generate a random sequence $(\sigma^{(t)})_{t=0,\ldots,T}$ of configurations by updating the status of a random sample at each time step according to \eqref{eqNishi}; see~\cite{Peres} for a detailed derivation of the Glauber update rule.
The hope is that for moderate $T$ the empirical means of the sequence approximate the actual posteriors well, i.e.,
\begin{align}\label{eqGlauber}
	\mu_{G}(\{\vsigma(x_j)=s\})&\approx\frac{1}{T}\sum_{i=0}^T\vecone\cbc{\sigma^{(t)}(x_j)=s}&&(j=1,\ldots,n;\,s=0,1).
\end{align}

At this point, no mathematical analysis of Glauber exists.
Furthermore, an empirical assessment of \eqref{eqGlauber} is difficult because even for moderate values of $n$ we cannot hope to compute the marginals of the posterior \eqref{eqNishi} exactly by exhaustive enumeration.
Nonetheless, an experimental study of Glauber has been conducted in~\cite{cuturi_2020}.

\subsection{Informative Dorfman}\label{Sec_informative}
Even if we assume that Glauber (or some other algorithm) approximates the posterior marginals well, how could we use this information in the second stage?
A simple idea is to revisit the original Dorfman design.
Hence, equipped with the posterior marginals from the first round, we could set up test pools such that each sample gets placed in precisely one pool.
But now we could try to take the posteriors from the first stage into consideration in compiling the pools.
Finally, just like in the original Dorfman scheme one could test the samples in each pool that returns a positive result separately.
This procedure goes by the name of {\em informative Dorfman} \cite{mcmahan_2012}.

How exactly do we take advantage of the marginals to set up the pools?
A natural idea is to sort the samples increasingly according to their marginals and pool them in this order.
A simple optimisation given the sequence of marginals then yields the optimal sequence of pool sizes.
The pools containing samples with small marginals are relatively large, while samples with marginals above $0.3$ get tested individually.
A combination of Glauber and informative Dorfman has been studied empirically in~\cite{cuturi_2020}. The key finding was that for a given number of tests this procedure worked decently well but was still outperformed by quite a margin by more complicated multi-stage test designs and algorithms.
In our study, we find that the marginals obtained by running Belief Propagation closely resemble the empirical marginals of Glauber and thus consistently use Belief Propagation in our analyses.

\section{Adaptive Belief Propagation} \label{sec_adabp}

\noindent
In this section we discuss the new design and inference algorithm.
The first stage employs the random biregular test design from \Sec~\ref{Sec_bireg}.
Given the results of the first stage, in the second and third stage we use a blend of the random biregular design and informative Dorfman.
For the inference algorithm we seize upon the Belief Propagation message passing paradigm \cite{Pearl}.
Since Belief Propagation and the mathematical theory behind this algorithm inform the entire construction, that is where we start.

\subsection{Belief Propagation}\label{Sec_BP}
In recent years the Belief Propagation message passing paradigm has been applied in combination with randomised constructions with stunning success.
Prominent examples include coding theory and other signal processing tasks such as compressed sensing~\cite{Donoho_2013,Krzakala_2012,RichardsonUrbanke}.
The development of the Belief Propagation technique in conjunction with randomised constructions has been inspired by deep ideas from the statistical mechanics of disordered systems~\cite{MM}.
More recently, a substantial body of mathematical research has been devoted to Belief Propagation; e.g.,~\cite{harnessing,Will,Charis,vontobel}.
Although most of the theoretical work from both the physics and maths communities is intrinsically asymptotical, we let these ideas guide our quest for a practical group testing design.

Belief Propagation is a generic message passing technique for approximating the marginals of Boltzmann distributions on factor graphs.
The posterior distribution \eqref{eqNishi} turns out to be a specimen of such a Boltzmann distribution.
The basic intuition behind Belief Propagation, which has been substantiated mathematically to a good extent, is that under certain assumptions the posterior distribution admits a succinct representation in terms of {\em messages}~\cite{Will,Will2,MM,Zdeborova_2016}.
These assumptions are provably met in many Bayes-optimal inference problems on random factor graphs, at least asymptotically as the problem size tends to infinity~\cite{Jean,CEJKK}.
The group testing problem as modelled in \Sec~\ref{Sec_model} is an example of such a Bayes-optimal inference problem.

At first glance the posterior distribution~\eqref{eqNishi} appears to be quite a difficult object of study.
For instance, if we were to estimate the entropy of this distribution we might have to inspect all $2^n$ possible vectors $\sigma\in\{0,1\}^n$.
But according to the Belief Propagation paradigm we can get a handle on the posterior distribution in terms of messages associated with the edges of the test design $G=G_{n,m}(\Gamma,\Delta)$.
Formally, the {\em message space} of $\cM(G)$ consists of vectors 
\begin{align*}
	(\mu_{x_j\to a_i}(s),\mu_{a_i\to x_j}(s))_{j=1,\ldots,n;\,i=1,\ldots,m;\,x_j\in\partial a_i;\,s\in\{0,1\}}.
\end{align*}
The idea is that there are two messages $\mu_{x_j\to a_i}(\nix)$, $\mu_{a_i\to x_j}(\nix)$ associated with every edge of $G$, one directed from the sample $x_j$ to the test $a_i$ and one in the opposite direction.
The messages themselves are probability distributions on $\{0,1\}$.
Thus, 
\begin{align*}
	\mu_{x_j\to a_i}(0),\mu_{x_j\to a_i}(1)\in[0,1]\quad\mbox{and}\quad\mu_{x_j\to a_i}(0)+\mu_{x_j\to a_i}(1)=1,
\end{align*}
and similarly for $\mu_{a_i\to x_j}(\nix)$.

Roughly speaking, $\mu_{a_i\to x_j}(\nix)$ is meant to represent the impact that $a_i$ has on $x_j$ in the absence of all other tests $b\in\partial x_j$.
Moreover, $\mu_{x_j\to a_i}(\nix)$ represents the status of $x_j$ in the absence of test $a_i$.
More formally, we define the {\em standard message} $\mu_{G,x_j\to a_i}(s)$ as the posterior probability that $\vsigma(x_j)=s$ given the test design $G-a_i$ obtained from $G$ by omitting test $a_i$ and given the test results $(\hvsigma(a_h))_{h\neq i}$.
In light of \eqref{eqNishi} we can write this probability out explicitly as
\begin{align*}
	\mu_{G,x_j\to a_i}(s)&\propto\sum_{\substack{\sigma\in\{0,1\}^\cX\\\sigma(x_j)=s}}
	\prod_{i=1}^n\lambda^{\sigma(x_i)}(1-\lambda)^{1-\sigma(x_i)}\prod_{i=1}^m \psi_{\hvsigma(a_i)}\bc{(\sigma_y)_{y\in \partial a_i}}
	&&(j=1,\ldots,n;\, i=1,\ldots,m;\, x_j\in\partial a_i;\,s\in\{0,1\}),
\end{align*}
with the $\propto$-sign hiding the normalisation to ensure that $\mu_{G,x_j\to a_i}(0)+\mu_{G,x_j\to a_i}(1)=1$.
Similarly, the standard message $\mu_{G,a_i\to x_j}(s)$ is defined as the posterior probability that $\vsigma(x_j)=s$ given the test design $G-(\partial x_j\setminus\{a_i\})$ obtained by removing all tests that $x_j$ takes part in except for $a_i$ and given the test results $\hvsigma(a_h)$ of all tests $a_h\not\in\partial x_j\setminus\{a_i\}$.

Conceived wisdom, vindicated mathematically for a broad family of inference problems, predicts that asymptotically these messages satisfy the following {\em Belief Propagation equations}~\cite{Jean,CEJKK,Will,Zdeborova_2016}:
\begin{align}
	\mu_{G,x\to a}(s)&\propto \lambda^s(1-\lambda)^{1-s}\prod_{b\in\partial x\setminus\cbc a}\mu_{G,b\to x}(s),\label{eqBP_simple1}\\
	\mu_{G,a\to x}(0)&\propto 1-q+(p+q-1)\prod_{y\in \partial a \setminus x}\mu_{G,y\to a}(0),&\mu_{G,a\to x}(1)&\propto 1-q&\mbox{ if }\hvsigma(a)=0, \label{eqBP_simple2}\\
	\mu_{G,a\to x}(0)&\propto q+(1-p-q)\prod_{y\in \partial a \setminus x}\mu_{G,y\to a}(0),&\mu_{G,a\to x}(1)&\propto q&\mbox{ if }\hvsigma(a)=1.
	\label{eqBP_simple3}
\end{align}
These equations express the notion that the random biregular design $G_{n,m}(\Gamma,\Delta)$ minimises dependencies between the test results.
Furthermore, we expect that the marginals of the posterior distribution can be well approximated in terms of the messages:
\begin{align}\label{eqBP_simple_marg}
	\mu_G(\{\vsigma(x_i)=s\})&\propto\lambda^s(1-\lambda)^{1-s}\prod_{b\in\partial x_i}\mu_{G,b\to x_i}(s).
\end{align}

Apart from the marginals, asymptotic results also suggest that the entropy of the posterior distribution can be approximated in terms of the messages~\cite{CEJKK,Will,MM}.
This approximation comes in terms of a functional called the {\em Bethe free energy}, defined as
\begin{align}\label{eqBFE}
	\cB_G&=\sum_{x \in \cX}\cB_{G,x} + \sum_{a \in \cA} \cB_{G,a} - \sum_{x \in \cX, a \in\partial x} \cB_{G,x, a}  \qquad \text{with} \\
	\cB_{G,x} &= \log \sum_{s \in \cbc{0,1}} \prod_{a \in \partial x} \mu_{G,a \to x}(s) \\
	\cB_{G,a}&=
	\begin{cases}
		\log \bc{1-q+(p+q-1)\prod_{x \in \partial a} \mu_{G,x \to a}(0)}&\mbox{ if }\hvsigma(a)=0\\
		\log \bc{q+(1-p-q)\prod_{x \in \partial a} \mu_{G,x \to a}(0)} &\mbox{ if }\hvsigma(a)=1
	\end{cases} \\
	\cB_{G,x,a} & = \log \sum_{s \in \cbc{0,1}} \mu_{G,x \to a}(s) \mu_{G,a \to x}(s).
\end{align}
The resulting approximation of the entropy reads 
\begin{align}\label{eqEntroy}
	\cH_G&=\cB_G -n\log\lambda+\sum_{i=1}^n\mu_G(\{\vsigma_x=0\})\log\frac\lambda{1-\lambda}\\
		 &\quad-\sum_{\substack{i=1\\\hvsigma(a_i)=0}}^m \frac{p\prod_{x\in\partial a_i}\mu_{G,x\to a_i}(0)}{1-q+(p+q-1)\prod_{x\in\partial a_i}\mu_{G,x\to a_i}(0)}\log p+\frac{(1-q)(1-\prod_{x\in\partial a_i}\mu_{G,x\to a_i}(0))}{1-q+(p+q-1)\prod_{x\in\partial a_i}\mu_{G,x\to a_i}(0)}\log(1-q)\nonumber\\
		 &\quad-\sum_{\substack{i=1\\\hvsigma(a_i)=1}}^m \frac{(1-p)\prod_{x\in\partial a_i}\mu_{G,x\to a_i}(0)}{q+(1-p-q)\prod_{x\in\partial a_i}\mu_{G,x\to a_i}(0)}\log (1-p)+\frac{q(1-\prod_{x\in\partial a_i}\mu_{G,x\to a_i}(0))}{q+(1-p-q)\prod_{x\in\partial a_i}\mu_{G,x\to a_i}(0)}\log q.\nonumber
\end{align}

Hence, in order to estimate the marginals and the entropy of the posterior we need to calculate the Belief Propagation messages.
A natural idea is to perform a fixed point iteration using the Belief Propagation equations~\eqref{eqBP_simple1}--\eqref{eqBP_simple3}.
Of course, the equations \eqref{eqBP_simple1}--\eqref{eqBP_simple3} may possess several solutions; they usually do~\cite{CEJKK,Zdeborova_2016}.
Whether or not the fixed point iteration homes in on the correct solution then depends on the initialisation.

While there is no generic recipe for choosing an appropriate initialisation $\mu^{(0)}\in\cM(G)$, two choices suggest themselves.
First, we could initialise the messages according to the prior $\lambda$, i.e.,
\begin{align}\label{eqInitPrior}
	\mu_{x_j\to a_i}^{(0)}(s)&=\lambda^s(1-\lambda)^{1-s}.
\end{align}
Second, we could initialise the messages in accordance with the ground truth, i.e.,
\begin{align}\label{eqInitTruth}
	\mu_{x_j\to a_i}^{(0)}(s)&=\vsigma(x_j).
\end{align}
The latter is not practically useful for the obvious reason.
But the analogy with other applications of Belief Propagation for inference problems suggests that {\em if} the fixed point iteration converges to the same solution to~\eqref{eqBP_simple1}--\eqref{eqBP_simple3} from the two initialisations \eqref{eqInitPrior} and \eqref{eqInitTruth}, then this solution actually is a good approximation to the correct messages.
Furthermore, whether or not \eqref{eqInitPrior} and~\eqref{eqInitTruth} yield the same solution we can try out experimentally.

One last but crucial point remains to be clarified: how precisely do we perform the fixed point iteration?
The textbook method is to perform message updates in parallel.
This means that, starting from the initialisation $(\mu^{(0)}_{x_j\to a_i})_{i,j}$, we compute all test-to-sample approximations $\mu^{(0)}_{a_i\to x_j}$ via \eqref{eqBP_simple2}--\eqref{eqBP_simple3}.
Then we use these together with \eqref{eqBP_simple1} to compute the next approximation $(\mu^{(1)}_{x_j\to a_i}(\nix))_{i,j}$ to all sample-to-test messages, and so forth.

\begin{figure}[ht]
	\centering
	\includegraphics[scale=0.32]{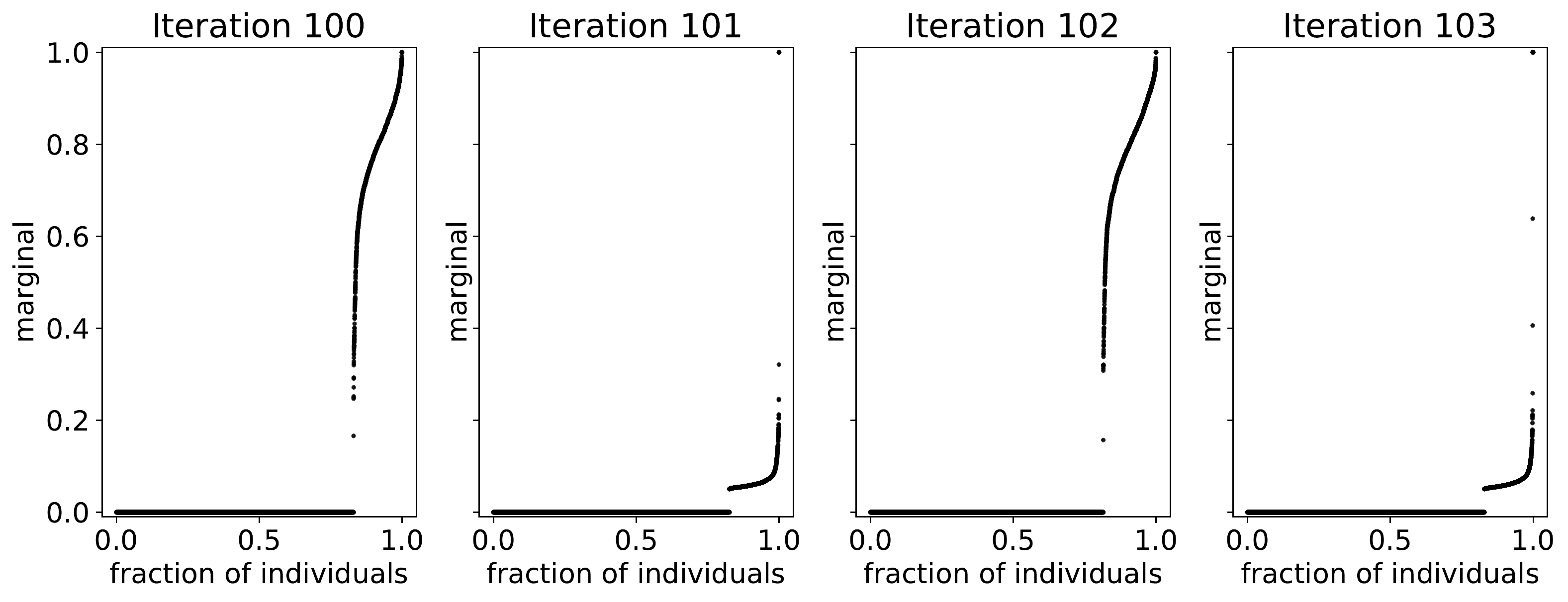}
	\caption{Illustration of the oscillatory behavior of Belief Propagation when performing parallel updates for $\lambda=0.05$ and $0.2$ tests/n for the noiseless setting}
	\label{fig_oscillation}
\end{figure}

This parallel updates mechanism was tried out experimentally in~\cite{cuturi_2020}.
However, this method does not converge.
Instead, the messages oscillate between odd and even rounds as shown in Figure~\ref{fig_oscillation}.
Similar oscillations emerge in other applications of Belief Propagation.
They may result from an instability of the empirical mean of the messages.
To elaborate, if in some particular iteration $t$ the deviation from the prior
\begin{align}\label{eqOscillate}
	\sum_{j=1}^n\sum_{i=1}^m\vecone\cbc{a_i\in\partial x_j}(\mu_{x_j\to a_i}^{(t)}(1)-\lambda)
\end{align}
is positive, then we should expect a negative deviation in the next round.
This is because due to \eqref{eqOscillate} in the next iteration many tests will receive a relatively large indication from that one of their samples may be infected.
The test will therefore send out ``less urgent'' messages to the other samples.
Conversely, if \eqref{eqOscillate} is negative, then in iteration $t+1$ we expect to see a positive deviation.
Due to the analytic nature of the update rules \eqref{eqBP_simple1}--\eqref{eqBP_simple3} these oscillations do not dampen down but actually blow up, leading to oscillations between odd and even rounds.
This observation led the authors of~\cite{cuturi_2020} to turn to the computationally more intensive Glauber dynamics algorithm.

But actually oscillations of this type have been observed in other problems as well and several ideas for tackling the problem are on the market.
Perhaps the most organic solution, and the method to which we resort, is to update the messages in a randomised fashion rather than in parallel.
Hence, starting from the initialisation $(\mu^{(0)}_{x_j\to a_i}(\nix))_{i,j}$, we apply \eqref{eqBP_simple2}--\eqref{eqBP_simple3} once to initialise the test-to-sample messages $\mu_{a_i\to x_j}(\nix)$ as well.
Then at each time $t\geq1$ we choose an edge $a_ix_j$ of $G$ randomly and also flip a fair coin.
If the coin comes up heads we update the message $\mu_{x_j\to a_i}^{(t)}(\nix)$ according to \eqref{eqBP_simple1}.
Otherwise we update $\mu_{a_i\to x_j}^{(t)}(\nix)$ according to \eqref{eqBP_simple2}--\eqref{eqBP_simple3}.
The random choices break the cycle of oscillations.
We stop the fixed point iteration after a fixed number $T$ of steps.
The precise choice of $T$ is guided by experiments but of course $T$ should be chosen large enough so that every message will likely get updated several times. We note that this update scheme does not impede practical matters from using our algorithm in a laboratory setting since it purely pertains to the computations behind the scene and does not impact how samples are split and combined. 

\begin{figure}[h]
	\centering
	\begin{minipage}[c]{0.49\textwidth}
		\centering
		\includegraphics[scale=0.3]{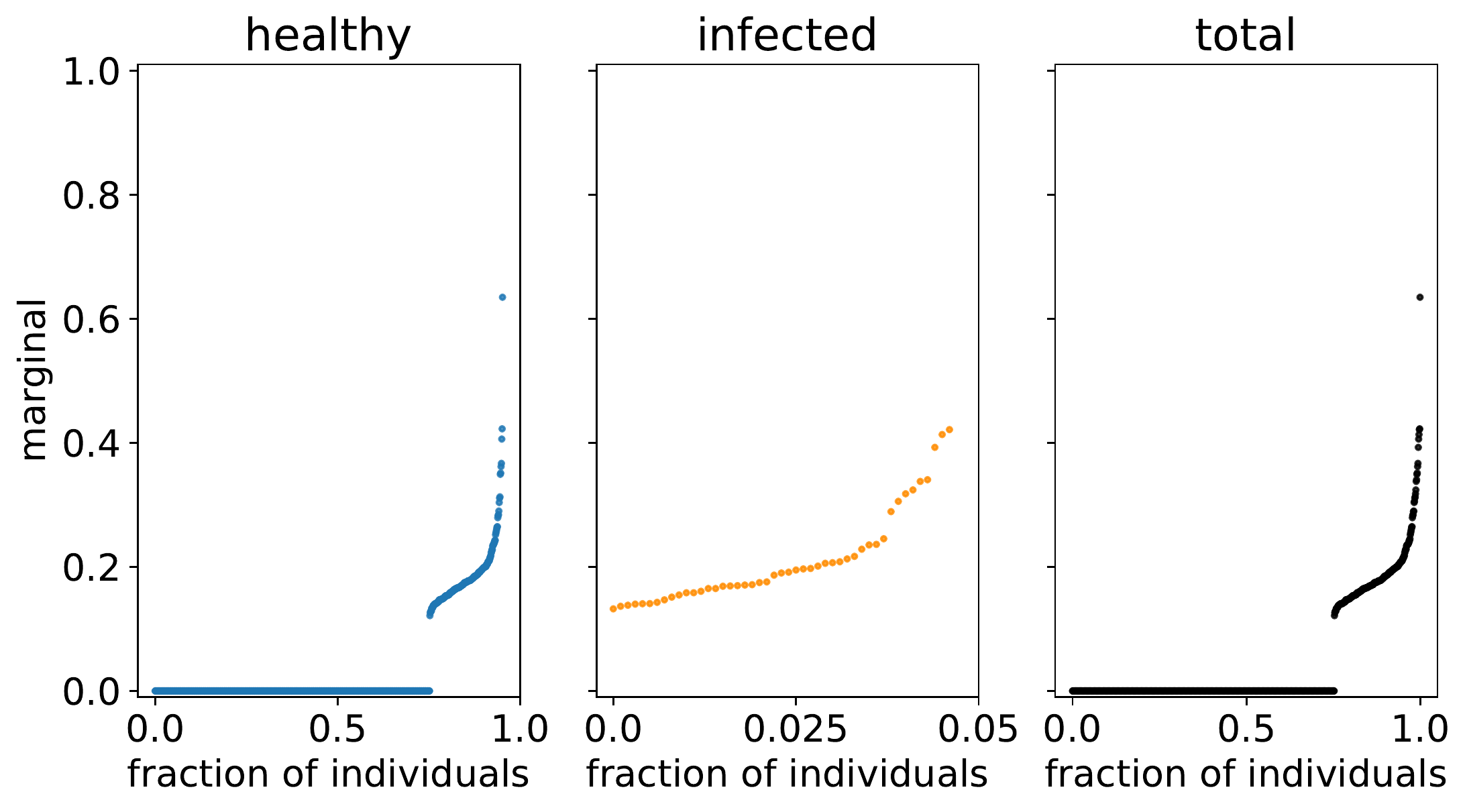}
	\end{minipage}
	\begin{minipage}[c]{0.49\textwidth}
		\centering
		\includegraphics[scale=0.3]{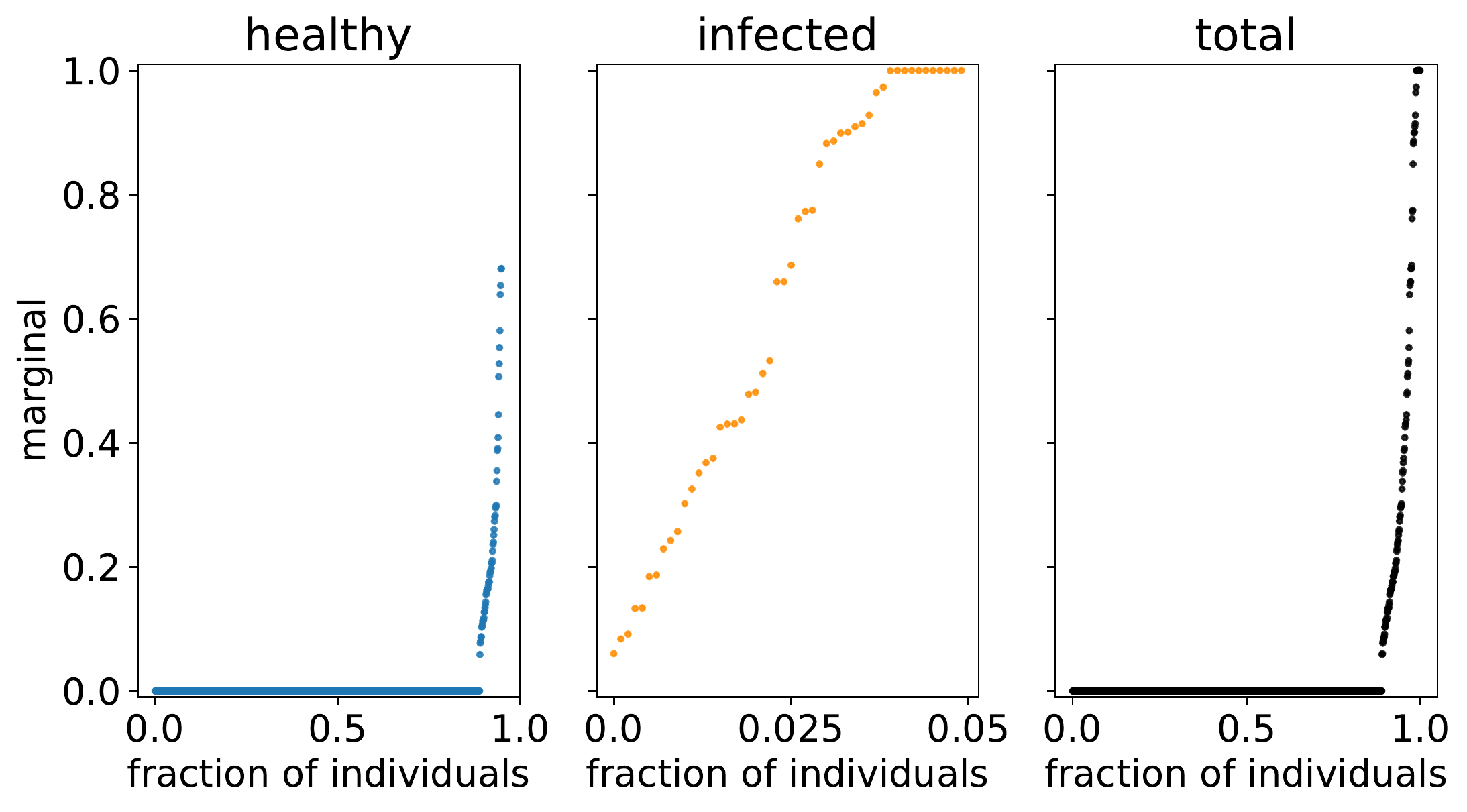}
	\end{minipage}
	\centering
	\begin{minipage}[c]{0.49\textwidth}
		\centering
		\includegraphics[scale=0.3]{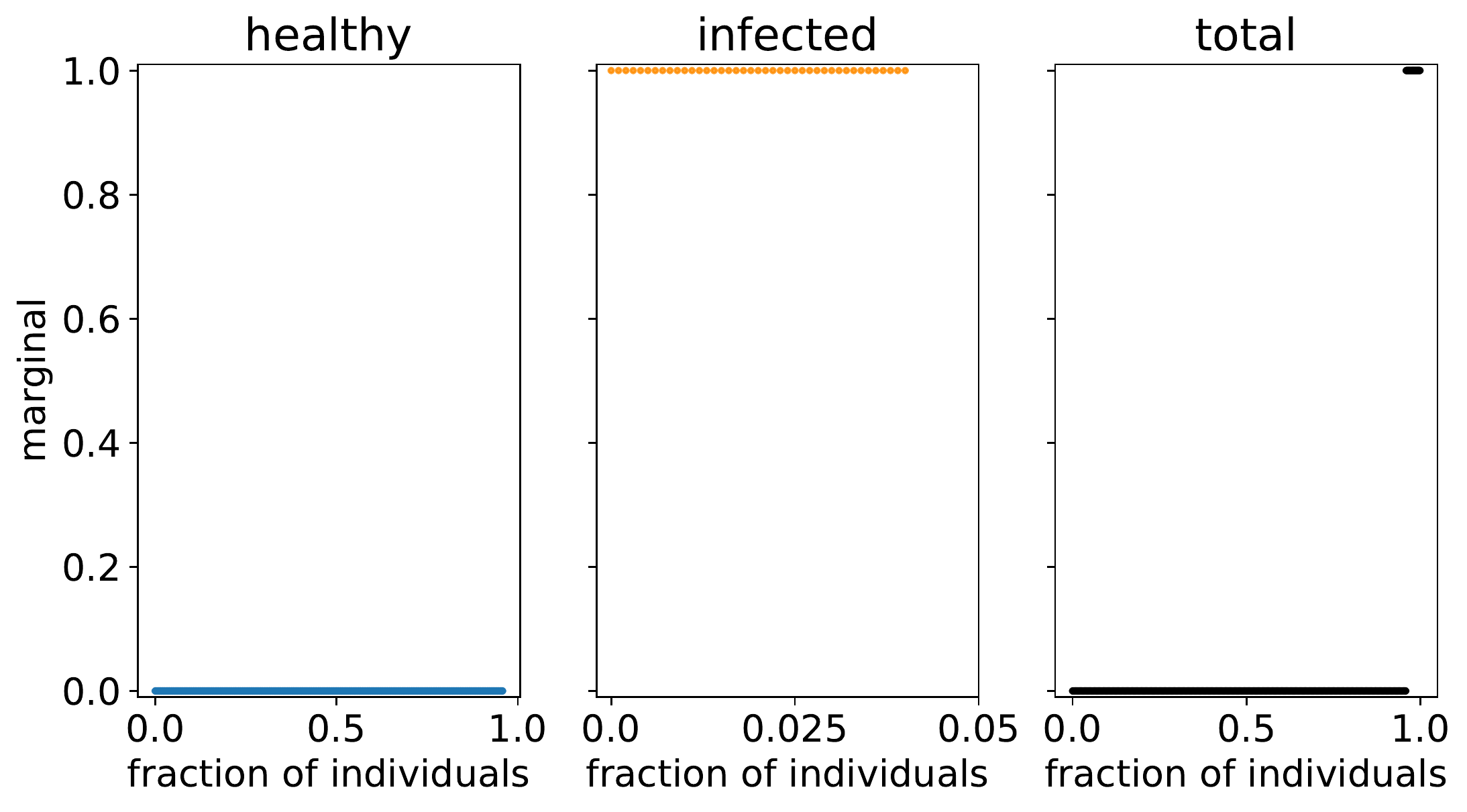}
	\end{minipage}
	\begin{minipage}[c]{0.49\textwidth}
		\centering
		\includegraphics[scale=0.4]{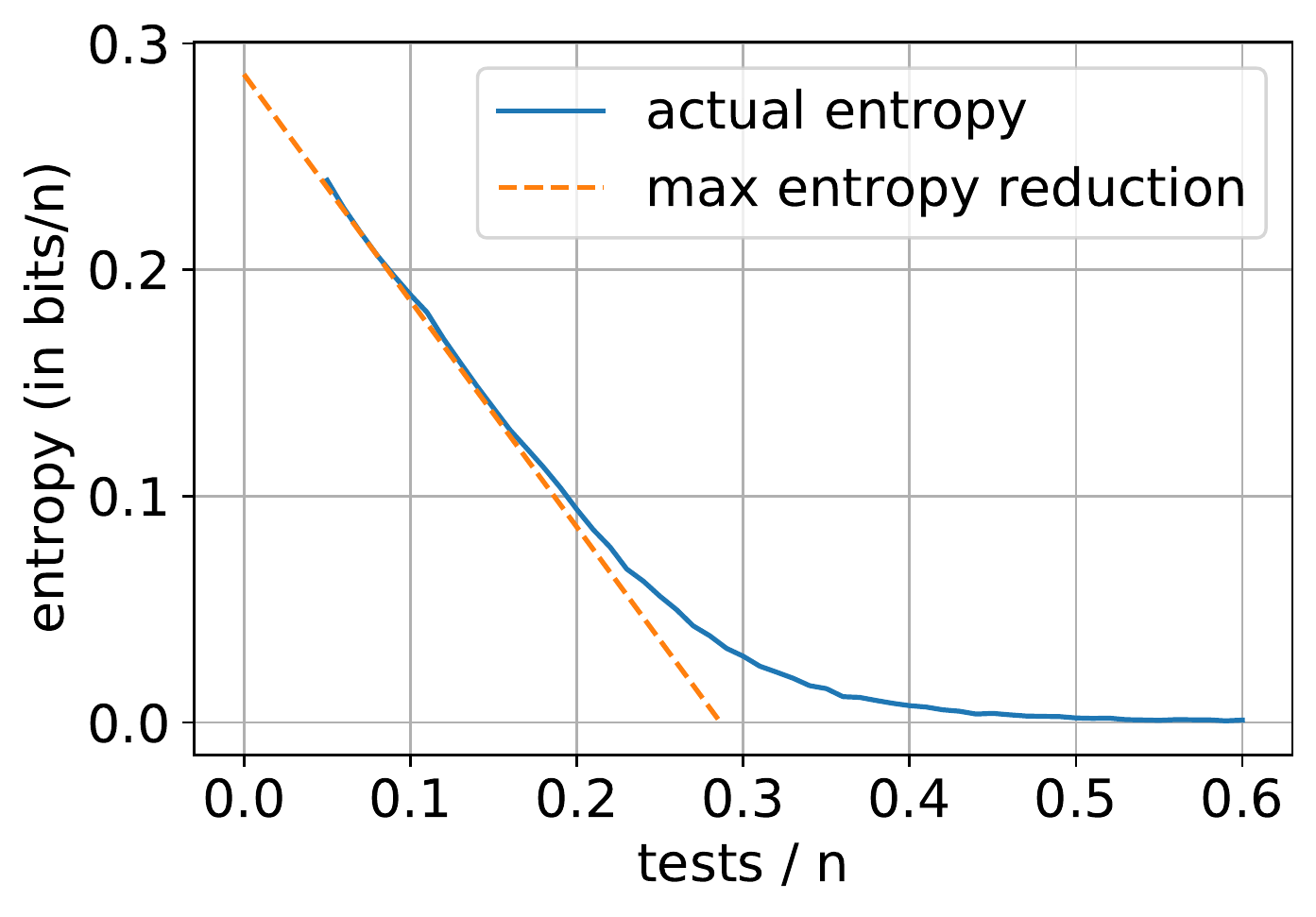}
	\end{minipage}
	\caption{Illustration of the posterior distribution from running Belief Propagation on a random biregular test design for with $0.15$ (top left), $0.25$ (top right) and $0.6$ (bottom left) tests/n and remaining entropy (bottom right) for $\lambda=0.05$ and the noiseless setting.}
	\label{fig_marginal}
\end{figure}

Beyond relying on asymptotic ideas and comparing the messages that result from the two aforementioned initialisations we take two additional steps to corroborate the results of Belief Propagation.
First, we compared the marginals obtained by Belief Propagation with the empirical marginals of Glauber dynamics on a number of samples.
They match.
Second, we compared the marginals obtained via Belief Propagation on moderately sized biregular test designs with the marginal distributions obtained via {\em population dynamics}, a heuristic intended to approximate the limiting distribution of the marginals as $n\to\infty$ \cite{MM}.
Once again the Belief Propagation results align very well.
Figure~\ref{fig_marginal} displays the typical outcome of the Belief Propagation for different numbers of tests along with the estimate \eqref{eqEntroy} of the remaining entropy.

What conclusions are to be drawn regarding a promising test design?
We see three different scenarios.
\begin{itemize}
\item For small numbers $m$ of tests we can extract some information from the negative tests.
	For instance, in the case $p=1$ of perfect specificity we can rest assured that any sample included in a negative test is indeed uninfected.
But beyond the direct effect of the negative tests the marginals do not align particularly well with the ground truth.
\item The second scenario concerns intermediate values of $m$.
	Here Belief Propagation gains information from both positive and negative test results.
	As a consequence, the marginals start to align better with the ground truth.
\item Finally, once $m$ gets quite large the ground truth leaves a clear imprint on the test results.
In this scenario we can recover the ground truth with good accuracy, albeit at the expense of investing many tests.
\end{itemize}
In light of what we learned on Belief Propagation, we now move on to describe the new \abp\ test design.

\subsection{The first stage}\label{Sec_stage1}
As the first stage we use the random biregular design $G=G_{n,m}(\Delta,\Gamma)$ with the optimal choice of $\Delta,\Gamma$ from \eqref{eqGammaDelta} subject to rounding.
Thus, the only free parameter is the total number $m$ of tests conducted in the first stage.
Its choice is informed by Belief Propagation.

Specifically, we choose the largest number $m$ of tests up to which each test yields the optimal entropy reduction of $\ln 2$.
Practically, this means that we choose $m$ to match the point at which the entropy plot for the corresponding parameter values flattens.
The fourth graphic in Figure~\ref{fig_marginal} shows the approximation of the entropy as a function of the number of tests for our illustrative case of $n=1000$ and $\lambda=0.05$ in the noiseless setting.
For other priors and noise levels, the story turns out to be analogous.

\subsection{The second and third stage}\label{Sec_stage2}

Given the approximation of the marginals from the first stage, how should we proceed?
As we saw in \Sec~\ref{Sec_bireg} and~\ref{Sec_informative}, two ideas for the subsequent stages proposed in the literature include individual testing of all samples whose marginals are not entirely polarised after the first round and informative Dorfman.
The former strategy, known as Definite Defectives, seems wasteful as it completely disregards any non-trivial information about the marginals resulting from the Belief Propagation computation.
The latter suffers from the same problem as the original Dorfman scheme, namely a potentially fairly large number of false positives and negatives.

\begin{figure}[h]
	\centering
	\includegraphics[scale=0.4]{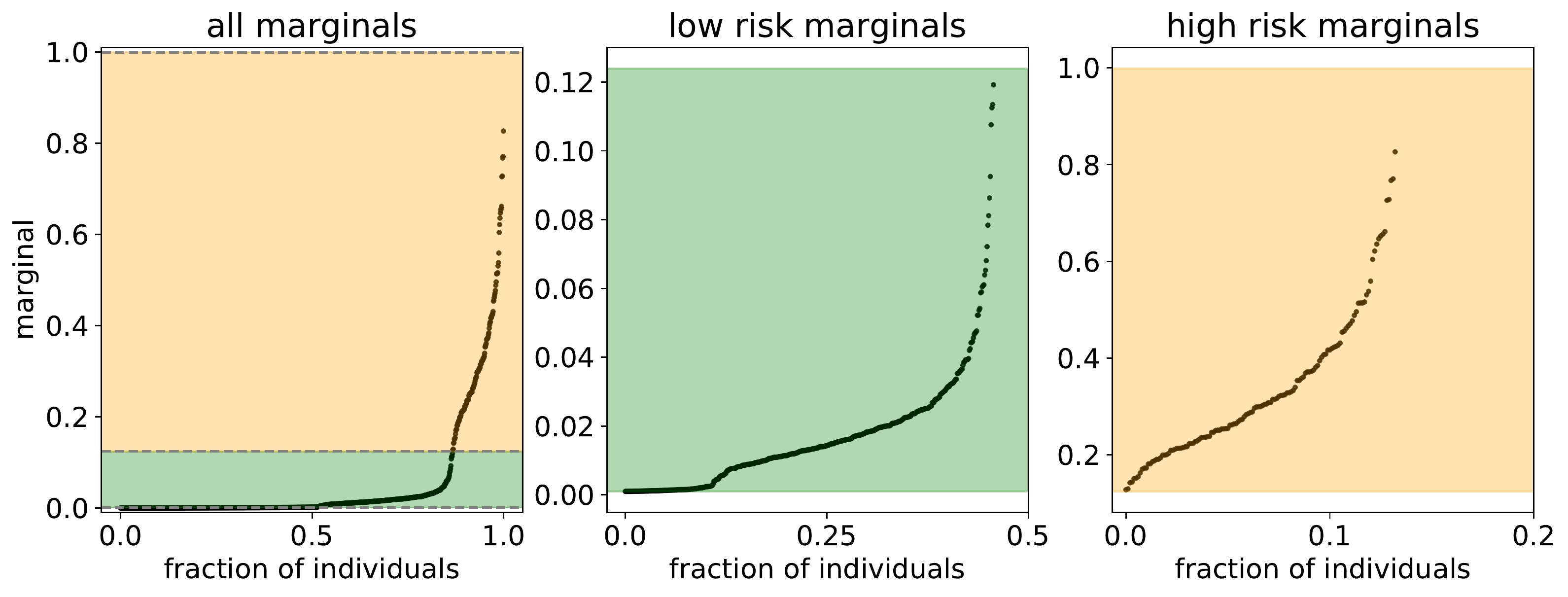}
	\caption{Illustration of split between low and high risk marginals for $\lambda=0.05$ in the high noise setting with $m/n=0.25$.}
	\label{fig_marginal_noise}
\end{figure}

To remedy these issues, we propose a new design that blends the random biregular design with the informative Dorfman scheme from \Sec~\ref{Sec_informative}.
For a start we threshold marginals obtained from the first stage at $0.1\%$ and $99.9\%$. 
Thus, we report samples with Belief Propagation marginals less than $0.1\%$ as healthy and those with marginals beyond $99.9\%$ as infected right after the first stage.
The remaining samples are split into two groups, one comprising samples with marginals below $12.4\%$ and one with marginals above. 
Let us refer to these as the {\em low risk} and the {\em high risk} groups, respectively.
The choice of $12.4\%$ marks precisely the threshold beyond which the expressions \eqref{eqGammaDelta} suggest that any sample should be placed in one test only. Figure~\ref{fig_marginal_noise} provides an illustration. 

For the low risk group we set up another random biregular test design on which we run Belief Propagation once again.
The posterior of the first stage now acts as the prior of the second stage. A range of different tests tested for the biregular test design depending on the prior, the posteriors of the first stage and noise level and the final recommended test numbers obtained via optimisation over this range.
The resulting marginals are again thresholded at $0.1\%$ and $99.9\%$. 
Those samples whose marginals fall in between are subsequently retested individually with their classification being solely determined by the outcome of the individual test.

To be more precise, let $\cX'$ be the samples in the low risk group, let $n'=|\cX'|$ and let $m'$ be the number of tests dedicated to this group.
Thanks to the Belief Propagation results from the first stage we can (approximately) calculate the average marginal
\begin{align*}
	\lambda'=\frac{1}{n'}\sum_{x\in\cX'}\mu_G(\{\vsigma(x)=1\}.
\end{align*}
Mimicing \eqref{eqGammaDelta} we then choose the degrees
\begin{align}\label{eqGammaDelta2}
	\Delta'&=\frac{m' \log(2)}{n' \lambda'}&\Gamma'&=\frac{\log(2)}{\lambda'}
\end{align}
subject to rounding and set up a random biregular test design $G'=G_{n',m'}(\Delta',\Gamma')$ on $\cX'$.
Furthermore, we modify the Belief Propagation equations on this random biregular design to accommodate the marginals computed in the first stage.
Hence, instead of using the universal prior $\lambda'$ for all the samples, we substitute the separate marginals computed in the first stage:
\begin{align}
	\mu_{G',x\to a}(s)&\propto \mu_{G}(\{\hvsigma(x)=1\})^s(1-\mu_{G}(\{\hvsigma(x))^{1-s}\prod_{b\in\partial x\setminus\cbc a}\mu_{b\to x}(s).\label{eqBP_second}
\end{align}
The test-to-sample equations remain the same as in \eqref{eqBP_simple2}--\eqref{eqBP_simple3}.

For the high risk group we set up an informative Dorfman design $G''$ as described in \Sec~\ref{Sec_informative}. If such a pooled test turns out to be negative, we classify all samples in this pool as healthy. Otherwise, we conduct individual tests and classify samples solely based on this individual test result.

\subsection{Enhanced accuracy}

The construction that we described up to this point is the one labelled \abp~1 in \Sec~\ref{Sec_results}.
Enhanced constructions \abp~2 and \abp~3 further reduce the number of false positives and negatives, at the expense of increasing the number of tests.
Indeed, the \abp~1 construction facilitates such enhancements explicitly. 
This is because almost all false positives and negatives actually originate from the informative Dorfman procedure in the second stage, while neither the thresholding nor the second-stage random biregular design tend to produce a notable number of mistakes.
Therefore, in \abp~2 and \abp~3 we simply perform the informative Dorfman procedure twice or thrice independently in parallel.
Thus, in \abp~2 and \abp~3 we double or triple the number of tests required for the informative Dorfman bit of the construction, {\em but only for that bit}.

If we perform informative Dorfman twice (\abp~2), we need to choose whether to reduce false negatives or false positives.
Accordingly, we classify a sample as healthy (infected) if both Dorfman procedures classify it as healthy (infected). 
In \abp~3 we get to avoid both false positives and false negatives.
To this end we classify according to the majority vote of the three informative Dorfman schemes. 
Table~\ref{tab_tests} illustrates the number of tests to be performed in the first and second stage depending on the prior and noise level.

\begin{table}
\begin{tabular}{ |p{4cm}|p{1cm}|p{1cm}|p{1cm}|p{1cm}|p{1cm}|p{1cm}|p{1cm}|  }
 \hline
 & & \multicolumn{2}{|c|}{noiseless} & \multicolumn{2}{|c|}{moderate noise} & \multicolumn{2}{|c|}{high noise} \\
 \hline
 algorithm & prior & m1/n & c & m1/n & c & m1/n & c \\
 \hline
 Belief Propagation + individual testing & 0.5\% & 0.05 & n/a & 0.09 & n/a & 0.11 & n/a \\
  & 1\% & 0.08 & n/a & 0.12 & n/a & 0.16 & n/a \\
  & 5\% & 0.23 & n/a & 0.37 & n/a & 0.45 & n/a \\
  & 10\% & 0.3 & n/a & 0.7 & n/a & 0.34 & n/a \\
  \hline
  Belief Propagation + informative Dorfman & 0.5\% & 0.045 & n/a & 0.05 & n/a & 0.045 & n/a \\
  & 1\% & 0.075 & n/a & 0.075 & n/a & 0.1 & n/a \\
  & 5\% & 0.28 & n/a & 0.24 & n/a & 0.16 & n/a \\
  & 10\% & 0.125 & n/a & 0.1 & n/a & 0.1 & n/a \\
  \hline
  \abp\ (1x) & 0.5\% & 0.035 & 1.0 & 0.05 & 2.0 & 0.05 & 2.0 \\
  & 1\% & 0.075 & 1.0 & 0.085 & 2.0 & 0.1 & 2.0 \\
  & 5\% & 0.28 & 1.0 & 0.18 & 2.0 & 0.16 & 2.0 \\
  & 10\% & 0.125 & 0.25 & 0.15 & 4.0 & 0.1 & 2.0 \\
  \hline
  \abp\ (2x) & 0.5\% & n/a & n/a & 0.075 & 8.0 & 0.02 & 8.0 \\
  & 1\% & n/a & n/a & 0.12 & 8.0 & 0.03 & 8.0 \\
  & 5\% & n/a & n/a & 0.4 & 2.0 & 0.36 & 2.0 \\
  & 10\% & n/a & n/a & 0.5 & 2.0 & 0.325 & 2.0 \\
  \hline
  \abp\ (3x) & 0.5\% & n/a & n/a & 0.075 & 8.0 & 0.02 & 8.0 \\
  & 1\% & n/a & n/a & 0.085 & 8.0 & 0.03 & 8.0 \\
  & 5\% & n/a & n/a & 0.4 & 2.0 & 0.4 & 2.0 \\
  & 10\% & n/a & n/a & 0.55 & 2.0 & 0.5 & 2.0 \\
  \hline
\end{tabular}
\label{tab_tests}
\caption{Number of tests for the first and second stage found via optimization for various algorithms, priors and noise levels. The number of tests in the second stage in terms of the stated parameter $c$ can be obtained as $c \lambda' n' \log(n')$ with $\lambda'$ and $n'$ defined as the average marginal and size of the low risk group, respectively.}
\end{table}

\section{Asymptotic considerations} \label{sec_asymptotic}
\noindent
Clearly,  \abp\ relies on heuristics and is not asymptotically optimal. 
This begs the question of how we would adapt the design and algorithm if we decide to live unburdened by practical considerations and consider the case $n \to \infty$? 

\subsection{Variations on \abp}
The optimal drop in entropy seen in Figure~\ref{fig_marginal} shows that running Belief Propagation on a random biregular test design in the first stage seems like a good idea. 
The discrete partition into three groups in the second stage, however, gives something away.
Indeed, in the asymptotic regime infinitesimal intervals of posterior marginals contain an unbounded number of samples.%
\footnote{Of course, depending on the prior and the noise setting the distribution of the posterior marginals need not be supported on the entire unit interval.} 
Thus, it seems information-theoretically optimal to construct a random biregular design for every single small marginal interval and repeat this procedure over a few stages. 
However, such an approach does not seem practical since for moderate $n$ each random biregular design would only contain very few samples. 

A simpler alternative that we considered is to still include all samples in one single second-stage test design, in which we choose the number of tests in which each sample takes part according to the posterior marginal from the first stage. 
Specifically, we chose these numbers so that in expectation half the tests should be positive.
However, this design turned out to be unstable for small values of $n$ because of random fluctuations. 

\subsection{Plain Belief Propagation}

Thus far we disregarded what might seem at first glance the most straightforward scheme: 
just run Belief Propagation on a random biregular design and then simply threshold the marginals at, say, $50\%$.
An obvious plus of this approach is that it requires one stage only.
Indeed, when we simulated this scheme for large group testing instances such as $n=10000$, this approach turned out to work extremely well.
Particularly for small priors such as $0.5\%$ and $1\%$, the plain Belief Propagation plus thresholding design is on par or even outperforms \abp\ in terms of both efficiency and reliability.
However, for smaller values of $n$ plain Belief Propagation plus threshold turns out to be extremely vulnerable to fluctuations of the number $\vk$ of infected samples.
This is because such fluctuations might cause the fraction of positive tests to significantly deviate from half.

\subsection{Scale effects} \label{sec_scale}

All the simulation results were presented for group testing instances with $n=1000$. However, we might be interested in smaller instances of say $n=100$ or larger ones such as $n=10000$. We performed extensive simulations in these directions and found that our results, particularly the power of the \abp\ scheme carry over to those group testing sizes as well, subject to rounding issues and few samples in a second stage for small instances necessitating slightly more tests. 

\subsection{Population dynamics}

\begin{figure}[h]
	\centering
	\begin{minipage}[c]{0.49\textwidth}
		\centering
		\includegraphics[scale=0.4]{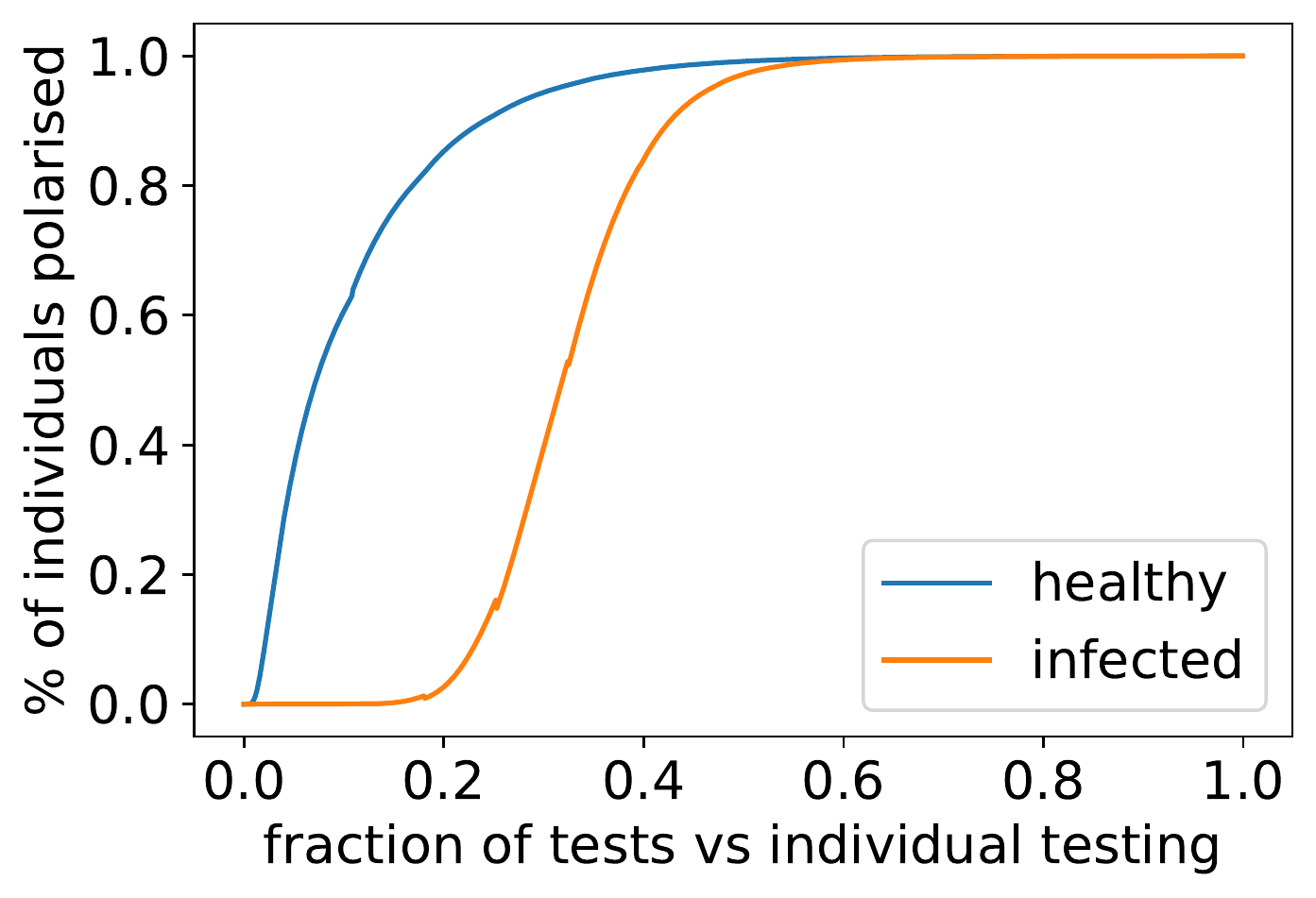}
	\end{minipage}
	\begin{minipage}[c]{0.49\textwidth}
		\centering
		\includegraphics[scale=0.4]{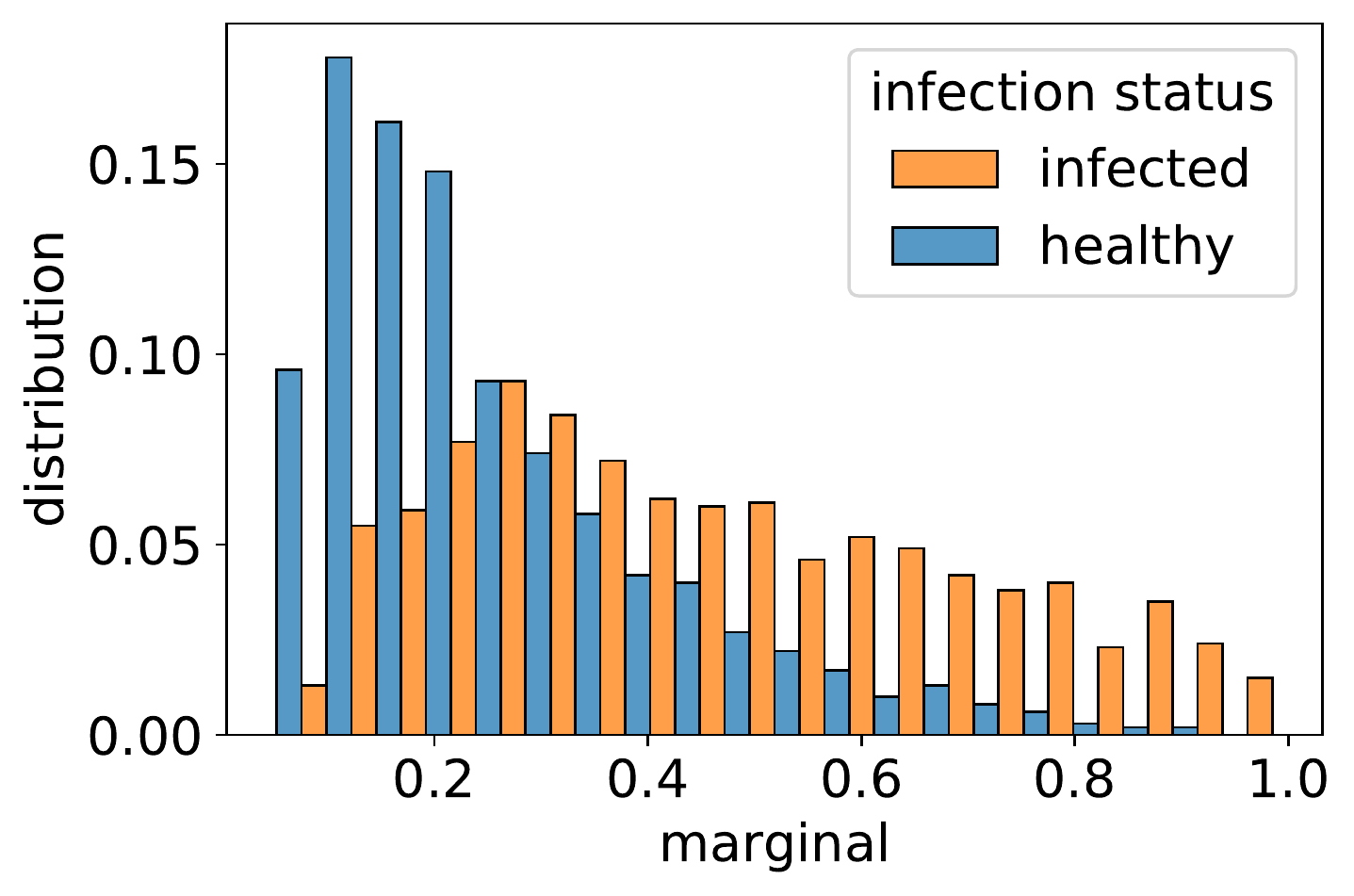}
	\end{minipage}
	\caption{Illustration of the asymptotic fraction of samples with polarised marginals and the posterior distribution for non-polarised samples obtained by running population dynamics on the offspring distribution by \cite{gebhard_2021} for $\lambda=0.05$ in the noiseless setting}
	\label{fig_pop_dyn}
\end{figure}

In the light of these scaling results for different instance sizes, let us spare a few more lines on the {\em population dynamics} already touched upon above. 
As mentioned above, this heuristic allows us to get a glimpse of the marginal distribution resulting from running Belief Propagation as $n \to \infty$~\cite{MM}. 
To this end, we require as input the distribution of infected and healthy samples in the local neighbourhood of a sample which is provided in \cite{gebhard_2021}. 
Subsequently, we iteratively sample the local neighbourhood for infected and healthy samples and perform one-step Belief Propagation updates to model the marginal distribution of those samples whose marginal is not completely polarised. 
The resulting distribution which is shown in Figure~\ref{fig_pop_dyn} for a prior of $5\%$ and the noiseless setting for illustration purposes closely resembles the marginal distribution that we observe from running Belief Propagation in our simulation in the first stage. 
As a side product, we obtain the proportion of polarised healthy and infected samples which lines up nicely with our simulation results. 
It should be noted that the population dynamics heuristic is nowhere near a complete analysis of Belief Propagation on random biregular graphs. 
Given the gains in efficiency and reliability that we observe in this empirical work for moderately-sized instances, a formal analysis of Belief Propagation seems to be an important next step in group testing research.

\section{Discussion}\label{sec_discussion}

\noindent
Group testing is a powerful method to efficiently and accurately detect infected samples.
Since the mathematical work on group testing deals with the asymptotic $n\to\infty$ scenario, practical adoption of methods proposed in this literature has been limited.
Instead practitioners tend to apply very simple test designs dating back to the 1940s.
In this paper we therefore conducted an experimental study that shows how a mildly more sophisticated test design can significantly improve the accuracy of the overall test results by comparison to classical methods without asking for many more tests.
The new test design comes with an efficient, easy-to-run and easy-to-implement algorithm that determines the status of each sample from the test results.
Since the new design employs randomisation, its adoption is probably feasible only in a practical setting that employs a degree of automation in preparing test pools.
But on the plus side the new \abp\ design keeps the pool sizes and the number of pools that each sample has to be placed in fairly low.

Apart from the group testing model studied in the present paper, there are several other, more complicated models.
For example, in {\em quantitative group testing} each test returns the {\em number} of infected samples rather than a binary positive or negative result.
Further variants include the pooled data problem, the generalised coin weighing problem or the compressed sensing problem \cite{donoho_2009,alaoui_2019}. 

What are the loose ends of the present work? 
On the one hand, it seems worthwhile to consider alternative noise models. 
A candidate might be one where the specificity decreases in the test size. 
Both the fixed noise model considered in this work and this diluted model have value from a practical perspective and it would be interesting to see whether our results carry over. 
On the other hand, the success of Belief Propagation in practical group testing leaves us wondering whether it is guaranteed to converge to a fixed point reminiscent of the ground truth. 
Hence, a mathematical analysis of Belief Propagation remains as an outstanding open problem.

\section*{Acknowledgement}
The authors thank Oliver Johnson for his helpful comments on group testing algorithms.

\bibliographystyle{siamplain}
\bibliography{references}

\begin{appendix}

\section{Sample splits and test degree}

The algorithms required the following number of maximum test degree and the following maximum and average split of samples. The algorithms can be readily adjusted to work with smaller test degrees or sample splits at the expense of slightly more tests.

\begin{tabular}{ |p{3cm}|p{1cm}|p{0.79cm}|p{0.79cm}|p{0.79cm}|p{0.79cm}|p{0.79cm}|p{0.79cm}|p{0.79cm}|p{0.79cm}|p{0.79cm}|  }
 \hline
 & & \multicolumn{3}{|c|}{noiseless} & \multicolumn{3}{|c|}{moderate noise} & \multicolumn{3}{|c|}{high noise} \\
 \hline
 algorithm & prior & $\Gamma_{\text{max}}$ & $\Delta_{\text{max}}$ & $\Delta_{\text{avg}}$ & $\Gamma_{\text{max}}$ & $\Delta_{\text{max}}$ & $\Delta_{\text{avg}}$ & $\Gamma_{\text{max}}$ & $\Delta_{\text{max}}$ & $\Delta_{\text{avg}}$ \\
 \hline
 \multirow{4}{3cm}{Belief Propagation + individual testing} & 0.5\% & 140 & 8 & 7.0 & 134 & 13 & 12.0 & 137 & 16 & 15.0 \\
 & 1\% & 75 & 7 & 6.0 & 67 & 9 & 8.0 & 69 & 12 & 11.0 \\
 & 5\% & 14 & 4 & 3.1 & 14 & 6 & 5.2 & 14 & 7 & 6.2 \\
 & 10\% & 7 & 3 & 2.3 & 8 & 6 & 5.2 & 6 & 3 & 2.8 \\
  \hline
  \multirow{4}{3cm}{Belief Propagation + informative Dorfman} & 0.5\% & 134 & 8 & 6.0 & 140 & 9 & 7.1 & 134 & 8 & 6.2 \\
 & 1\% & 67 & 7 & 5.1 & 67 & 7 & 5.2 & 70 & 9 & 7.2 \\
 & 5\% & 15 & 6 & 4.1 & 20 & 5 & 3.5 & 13 & 4 & 2.9 \\
 & 10\% & 8 & 3 & 1.8 & 14 & 3 & 2.3 & 10 & 3 & 2.3 \\
  \hline
  \multirow{4}{3cm}{\abp} & 0.5\% & 143 & 8 & 5.2 & 140 & 11 & 7.6 & 140 & 13 & 8.0 \\
 & 1\% & 67 & 8 & 5.1 & 71 & 12 & 6.7 & 70 & 13 & 8.0 \\
 & 5\% & 15 & 7 & 4.1 & 147 & 12 & 6.2 & 66 & 12 & 6.5 \\
 & 10\% & 8 & 3 & 1.8 & 172 & 19 & 10.3 & 50 & 10 & 4.9 \\
  \hline
\end{tabular}

\section{Distribution between stages}

Based on the number of tests in the first and second stage, the following table shows the fraction of samples identified in each round. It evinces that despite a total of three stages needed for \abp\ the majority of samples are identified already in the first and second stage, depending on the prior and noise level.

\begin{figure}[ht]
    \centering
    \includegraphics[scale=0.4]{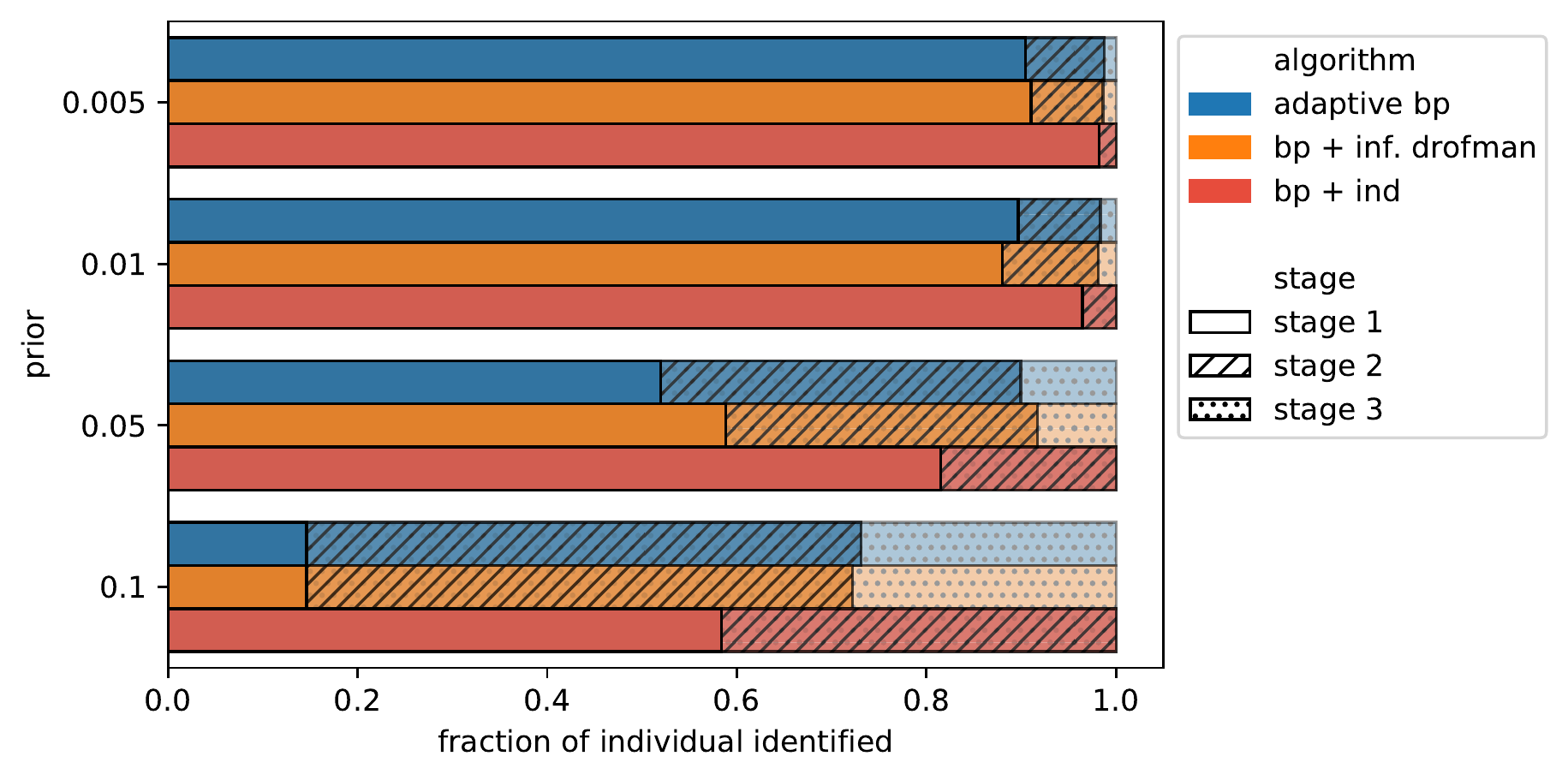}
    \caption{Fraction of samples identified in each stage by Belief Propagation followed by individual testing, Belief Propagation followed by informative Dorfman and \abp}
    \label{fig_stage_idst}
\end{figure}

\end{appendix}
\end{document}